\documentclass[11pt]{article}
\usepackage{jair}

\usepackage{theapa}
\usepackage{soul} 
\usepackage{amssymb}
\RequirePackage{ifthen}
\usepackage{amsmath}
\usepackage{latexsym}
\usepackage{qtree}
\usepackage{lscape}

\newlength\pseudocodewidth
\jairheading{28}{2007}{393-429}{6/06}{3/07}
\ShortHeadings{Bin Completion Algorithms for Multicontainer Problems}
{Fukunaga \& Korf}
\firstpageno{393}

\begin{document}

\title{Bin Completion Algorithms for Multicontainer Packing, Knapsack, and Covering Problems}

\author{\name Alex S. Fukunaga \email fukunaga@aig.jpl.nasa.gov\\
        \addr Jet Propulsion Laboratory \\
	      California Institute of Technology \\
              4800 Oak Grove Drive\\
              Pasadena, CA 91108 USA
         \AND
	 \name Richard E. Korf \email korf@cs.ucla.edu\\
	 \addr Computer Science Department \\
	 University of California, Los Angeles\\
	 Los Angeles, CA 90095}
\maketitle

\begin{abstract}
Many combinatorial optimization problems such as the bin packing and multiple knapsack problems involve assigning a set of
discrete objects to multiple containers. These problems can be used to model task and resource allocation problems in multi-agent systems and distributed systms, and can also be found as subproblems of scheduling problems.
We propose bin completion, a 
branch-and-bound strategy for one-dimensional, multicontainer packing problems.  Bin completion
combines a bin-oriented search space with a powerful dominance
criterion that enables us to prune much of the space. 
The performance of the basic bin completion framework can be enhanced by using
a number of extensions, including nogood-based pruning techniques 
that allow further
exploitation of the dominance criterion.  Bin completion is applied to
four problems: multiple knapsack, bin covering, min-cost covering, and
bin packing.  We show that our bin completion algorithms yield new,
state-of-the-art results for the multiple knapsack, bin covering, and
min-cost covering problems, outperforming previous algorithms by several orders of magnitude with respect to runtime on some classes of hard, random problem instances.
For the bin packing problem, we demonstrate
significant improvements compared to most previous results, but show that
bin completion is not competitive with current state-of-the-art
cutting-stock based approaches.
\end{abstract}


\newtheorem{definition}{Definition}
\newtheorem{theorem}{Theorem}
\newtheorem{dominance}{Dominance Criterion}
\newtheorem{proposition}{Proposition}

\section{Introduction}
\label{sec:introduction}

Many NP-hard problems involve assigning some set of discrete objects
to multiple containers.  In one class of problems, the objective is
to {\em pack} some items into a set of containers without exceeding the containers' capacities.
In a related class of problems, the goal is to {\em
cover} a set of containers by filling them up to at least some minimal
level (quota) using a set of items.  When both the containers and the items
are modeled as one-dimensional objects (possibly with an associated
cost/value function), we refer collectively to these problems as {\em
one-dimensional, multicontainer packing and covering problems}, or simply {\em
multicontainer packing problems}.  

One example of a multicontainer packing problem
is the {\em bin packing problem}: Given a set of items (numbers), and
a fixed bin capacity, assign each item to a bin so that the sum of the
items assigned to each bin does not exceed the bin capacity, and the number of bins used is minimized.  For
example, given the set of items 6, 12, 15, 40, 43, 82, and a bin
capacity of 100, we can assign 6, 12, and 82 to one bin, and 15, 40,
and 43, to another, for a total of two bins. This is an optimal
solution to this instance, since the sum of all the items, 198, is
greater than 100, and hence at least two bins are required.


Multicontainer packing problems are ubiquitous.  They model many
important operations research problems such as cargo
loading and transport, and also model many artificial intelligence
applications, such as the allocation and rationing of resources or tasks
among a group of agents.   
Multicontainer packing problems can
often be found embedded as subproblems of more complex, real-world
combinatorial optimization problems. For example many constraint 
programming problems contain ``bin packing'' or ``knapsack''
constraints as subproblems \cite<e.g.,>{Shaw04}. Such constraints are at the core of many scheduling and resource allocation problems.

In this paper, we propose {\em bin completion}, a new algorithm for optimally solving 
multicontainer packing problems.
We begin in Section \ref{sec:problems} with an overview of four representative, strongly NP-complete, multicontainer problems:
(1) the bin packing problem, (2) the multiple knapsack
problem, (3) the bin covering problem, and (4) the min-cost covering
problem. 

In Section
\ref{sec:bin-completion}, we begin by describing the standard,
``item-oriented'' branch-and-bound framework for these problems.
In this traditional approach, items are
considered one at a time. Each node in the search corresponds to a
decision regarding the assignment of an item to some non-full container. 
Then, we describe {\em bin completion}, an alternative,
``bin-oriented'' branch-and-bound strategy with two key features: (1)
the nodes in the search tree correspond to complete assignments of items to a
single bin, and (2) {\em dominance criteria} between assignments of items to bins are used
to prune the search. 
Section \ref{sec:extensions} describes extensions to the
basic bin completion framework that improve search efficiency, as well as the runtime and memory usage at each node.

In Sections \ref{sec:multiple-knapsack}-\ref{sec:bin-packing}, we
explore the application of the bin completion framework to four
specific, one-dimensional multicontainer packing problems.  For each problem, we
review the previous work on algorithms that optimally solve the problem,
detail our bin completion algorithm for the problem, and provide an
empirical comparison with the previous state-of-the-art algorithm.
We apply bin completion to the multiple knapsack problem in Section
\ref{sec:multiple-knapsack}, and show that our bin completion
solver significantly outperforms Mulknap \cite{Pisinger99},
the previous state-of-the-art algorithm.
The min-cost covering problem (also called the liquid loading problem)
was the first problem for which \citeA{ChristofidesMT79}
proposed an
early variant of the bin completion approach.  In Section
\ref{sec:min-cost-covering}, we 
show that our new bin completion algorithm significantly outperforms
the earlier algorithm by Christofides et al.
In Section \ref{sec:bin-covering}, we apply bin completion to the bin covering problem (also known as the dual bin packing problem).
We show that our bin completion
algorithm significantly outperforms the previous state-of-the-art
algorithm by Labb\'{e}, Laporte, and Martello \citeyear{LabbeLM95}.
In Section \ref{sec:bin-packing}, we 
apply our extended bin completion algorithm to the bin packing problem. Although our initial results were promising \cite{Korf02,Korf03}, 
our best bin completion solver is not
competitive with the current state of the art, which is a recent
branch-and-price approach based on a cutting-stock
problem formulation.
Section \ref{sec:conclusions} concludes with a discussion and directions for future work.

\subsection{One-Dimensional, Multicontainer Packing Problems}
\label{sec:problems}

We consider the class of {\em multicontainer packing problems}:
Given a set of {\em items}, which must be assigned to 
one or
more {\em containers} (``bins''), each item can be assigned to at most one container.
Each item $j$ has a {\em weight} $w_j$ associated with it.
Depending on the problem, each item $j$ may also have 
a  {\em profit} or {\em cost}
$p_j$ associated with it.
We assume that item weights and containers are one-dimensional.  For
some real-world applications, such as continuous call double
auctions \cite{KalagnanamDL01}, this model can be applied directly.  For other
applications, such as the lot-to-order matching problem, the one-dimensional model is an approximation \cite{CarlyleKF01}.
We consider two types of containers:
(1) Containers with a {\em capacity}, where the
sum of the weights of the items assigned to a container cannot exceed
its capacity, and (2) containers with a {\em quota}, 
where the sum of the weights of the items assigned to
a container must be at least as large as the quota.  
When there is only a {\em single} container,
we have the well-known {\em 0-1 knapsack problem}.
See the recent text by Kellerer, Pferschy, and Pisinger
\citeyear{KellererPP04} for an overview of work on the 0-1 Knapsack
problem and its variants.
In this paper, we focus on four one-dimensional, multicontainer packing problems (1)
bin packing, (2) multiple knapsack, (3) bin covering, and 
(4) min-cost covering.

\subsubsection{The Bin Packing Problem}

In the {\em bin packing problem},
the goal is to pack $n$ items
with weights $w_1,...,w_n$ into bins of capacity $c$ such that all
items are packed into the fewest number of bins, and the sum of the
weights of the items in each bin is no greater than the capacity.
Classical applications of bin packing include the classic vehicle/container loading problem \cite{EilonC71}, as well as memory/storage allocation for data.
The minimal number of agents required to carry out a set of tasks in a
multiagent planning problem can be modeled as a bin packing problem.

More formally, the bin packing problem can be formulated as the integer program:

\begin{align} \label{eqn:bin-packing-first}
\mbox{minimize} &\sum_{i=1}^{n} y_i \\
\mbox{subject to:}
\label{eqn:bin-packing-capacity} 
&\sum_{j=1}^{n} w_j x_{ij} \leq c y_i,   &i=1,...,n \\
\label{eqn:bin-packing-occurrence}
&\sum_{i=1}^{n} x_{ij} \leq 1,  &j=1,...,n\\
&x_{ij} \in \{0,1\} & i=1,...,n , j = 1,...,n\\ 
\label{eqn:bin-packing-last}
&y_{i} \in \{0,1\}, & i=1,...,n 
\end{align}

\noindent where $y_i$ represents whether the $i$th bin is used 
or not ($y_i=1$ if any items are assigned to bin $i$, $y_i=0$ otherwise),
and $x_{ij}$ = 1 if item $j$ is assigned to bin $i$, and 0 otherwise.
Constraint \ref{eqn:bin-packing-capacity} ensures that the capacity is
not violated for each bin that is instantiated, and constraint
\ref{eqn:bin-packing-occurrence} ensures that items are assigned to at most one bin.

In this standard formulation, we assume that all bins have the same capacity.
However, this assumption is not restrictive, since 
instances where bins with different capacities
(non-uniform bins) can be 
modeled by introducing additional items and constraints.

\subsubsection{The 0-1 Multiple Knapsack Problem}
\label{sec:mkp-definition}

Consider $m$
containers with capacities $c_1,...,c_m$, and a set of $n$ items,
 where each item has a weight $w_1,...,w_n$ and profit $p_1,...,p_n$.
Packing the items in the containers to maximize the total profit of
the items, such that the sum of the item weights in each container does
not exceed the container's capacity, and each item is assigned to at most one container is the {\em 0-1 Multiple Knapsack
Problem}, or MKP.  

The MKP is a natural generalization of the 0-1
Knapsack Problem where there are $m$ containers of capacities $c_1,
c_2, ...c_m$.  Let the binary decision variable $x_{ij}$ be 1 if item
$j$ is placed in container $i$, and 0 otherwise.
Then the 0-1 Multiple Knapsack Problem can be formulated as:
\begin{align} \label{eqn:mkp-first}
\mbox{maximize} &\sum_{i=1}^{m}\sum_{j=1}^{n}p_{j}x_{ij}  \\ 
\mbox{subject to:}
\label{eqn:mkp-capacity}
&   \sum_{j=1}^{n}w_{j}x_{ij} \leq c_i , &i=1,...,m  \\
\label{eqn:mkp-single}
&   \sum_{i=1}^{m} x_{ij} \leq 1, &j=1,...,n \\
&    x_{ij}\in \{0, 1\} &\forall i, j. \label{eqn:mkp-last} 
\end{align}

Constraint \ref{eqn:mkp-capacity} encodes the capacity constraint for each container, and constraint \ref{eqn:mkp-single} ensures that each item is assigned to at most one container.

The MKP has numerous applications, including 
task allocation among a group of autonomous agents in order to maximize the total utility of the tasks executed
\cite{Fukunaga05}, continuous double-call auctions
\cite{KalagnanamDL01}, 
multiprocessor scheduling \cite{LabbeLM03}, vehicle/container loading
\cite{EilonC71}, 
and
the assignment of files to storage devices in order to maximize the
number of files stored in the fastest storage devices
\cite{LabbeLM03}.
A special case of the MKP that has been studied in its own right is the {\em Multiple Subset-Sum Problem} (MSSP), where the profits of the items are equal
to their weights, i.e., $p_j = w_j$ for all $j$
(e.g., \citeR{CapraraKP00} \citeyearR{CapraraKP00a},\citeyearR{CapraraKP03}).
An application of the 
MSSP is the {\em marble cutting problem}, where given $m$ marble
slabs, the problem is to decide how to cut the slabs into sub-slabs
(each sub-slab is then further processed into a product) in order to
minimize the total amount of wasted marble.

\subsubsection{Bin Covering}
\label{sec:bc-definition}

Suppose we have $n$ items with weights $w_1,...,w_n$, and an infinite
supply of identical containers with {\em quota}  $q$.
The {\em bin covering} problem, also known as the {\em dual bin
packing} problem is to pack the items into containers such that the
number of containers that contain sets of items whose sums are at
least $q$ is {\bf maximized}. That is, the goal is to distribute, or
ration the items among as many containers as possible, given that the
containers have a specified quota that must be satisfied.  Note that
the total weight of the items placed in a container can be greater
than $q$ (we assume infinite capacity, although assigning an additional item to a bin whose quota is already satisfies is clearly suboptimal).

More formally, the bin covering problem can be formulated as the integer program:

\begin{align} \label{eqn:bin-covering-first}
\mbox{maximize} &\sum_{i=1}^{n} y_i \\
\mbox{subject to:}
\label{eqn:bin-covering-quota} 
&\sum_{j=1}^{n} w_j x_{ij} \geq q y_i,   &i=1,...,n \\
\label{eqn:bin-covering-occurrence}
&\sum_{i=1}^{n} x_{ij} \leq 1,  &j=1,...,n\\
&x_{ij} \in \{0,1\} & i=1,...,n , j = 1,...,n\\ 
\label{eqn:bin-covering-last}
&y_{i} \in \{0,1\}, & i=1,...,n 
\end{align}

where $y_i$ represents whether the quota on the $i$th bin is satisfied ($y_i=1$)
or not ($y_i=0$), and $x_{ij}$ = 1 if item $j$ is assigned to bin $i$, and 0 otherwise.
Constraint \ref{eqn:bin-covering-quota} ensures that the quota is
satisfied for each bin that is instantiated, and constraint
\ref{eqn:bin-covering-occurrence} ensures that items are assigned to at most one bin.

Bin covering is a natural model for resource or task allocation among multiple agents where the goal is to maximize the number of agents who achieve some quota.
It is also models
industrial problems such as: (1) packing peach
slices into cans so that each can contains at least its advertised net
weight in peaches, and (2) breaking up monopolies into smaller
companies, each of which is large enough to be viable
\cite{AssmannJKL84}.
Another application of bin covering is the lot-to-order matching problem in the semiconductor industry, where the problem is to assign fabrication wafer lots to customer orders of various sizes \cite{CarlyleKF01}.

\subsubsection{Min-Cost Covering Problem (Liquid Loading Problem)}
\label{sec:mccp-definition}
We define the
{\em Min-Cost Covering Problem} (MCCP) as follows.
Given a set of $m$ bins with quotas $\{q_1,...,q_m\}$, and a set of
$n$ items with weights $w_1,...,w_n$ and costs $p_1,...,p_n$,
assign some subset of the
items to each bin such that (1) each item is assigned to at most
one bin, (2) the sum of the weights of the items assigned to each bin is
at least the bin's quota (i.e., the bin is covered, as in bin
covering), and (3) the total cost of all the items that are assigned to
a bin is minimized.  
This problem has also been called the {\em
liquid loading problem} \cite{ChristofidesMT79}, because it was
originally motivated by the following application: Consider the
disposal or transportation of $m$ different liquids (e.g., chemicals)
that cannot be mixed. If we are given $n$ tanks of various sizes,
each with some associated cost, the problem is to load the $m$ liquids
into some subset of the tanks so as to minimize the total cost.
Note that here, the ``liquids'' correspond to
containers, and the ``tanks'' correspond to the items.

Other applications of the MCCP include: (1) the
storage of different varieties of grain in different silos, where
different types of grains cannot be mixed, (2) storage of
food types in freezer compartments, (3) a trucking firm which
distributes its trucks (of different sizes) among customers with no
mixing of different customer's orders on any truck, and (4) storage of
information in storage devices (e.g., sensitive customer data that
must be segregated between physical filing cabinets or file servers). 
A closely related problem is the segregated storage problem \cite{Neebe87,EvansT93}.

More formally, the MCCP can be formulated as the integer program:

\begin{align} \label{eqn:mccp-first}
\mbox{minimize} &\sum_{i=1}^{m} \sum_{j=1}^{n} p_j x_{ij} \\
\mbox{subject to:}
\label{eqn:mccp-quota}
&\sum_{j=1}^{n} w_j x_{ij} \geq q_i,   &i=1,...,m\\
\label{eqn:mccp-single}
&\sum_{i=1}^{m} x_{ij} \leq 1,   & j=1,...,n\\
&x_{ij} \in \{0,1\}  &\forall i,j. \label{eqn:mccp-last}
\end{align}

The binary variable $x_{ij}$ represents whether item $j$ is assigned to container $i$.
Constraint \ref{eqn:mccp-quota} ensures that the quotas of all the bins
are satisfied, and constraint \ref{eqn:mccp-single} ensures that each item is assigned to at most one bin.
%

\subsubsection{A taxonomy of multicontainer problems}

Table \ref{tab:multi-container-problems} summarizes the two key, defining
dimensions of the four multicontainer problems we study in this
paper. One key dimension is whether the problem involves packing a set
of items into containers so that the container capacity is not
exceeded (packing), or whether the problem requires satisfying the
quota associated with a container (covering). Another key dimension is
whether all of the items must be assigned to bins, or whether only a
subset of items are selected to be assigned to bins. 
A third dimension (not shown in the table) is the set of item {\em
attributes} (e.g., weight, profit/cost). Bin packing and bin covering
are single-attribute problems (items have weight only), while the
multiple knapsack and min-cost covering problems are two-attribute
problems (items have weight and profit/cost). 

We focus on the bin packing, bin covering, MKP, and MCCP problems
because we believe that these are in some sense the most ``basic''
multicontainer problems. Many combinatorial optimization problems can
be viewed as extensions of these problems with additional
constraints. For example, the generalized assignment problem can be
considered a generalization of a MKP with a more complex profit
function.

\begin{table}[b]
  \begin{center}
    \begin{tabular} { | l | l | l | l | }
      \hline
      & assign {\bf all} items to bins & assign {\bf subset} of items to bins\\ \hline
      packing & bin packing & multiple knapsack \\ \hline
      covering & bin covering & min-cost covering \\
      \hline
    \end{tabular}
  \end{center}
\caption {Characterizing multicontainer problems.}
\label{tab:multi-container-problems}
\end{table}

\section{Bin Completion}
\label{sec:bin-completion}

The multiple-knapsack problem, bin covering problem, min-cost covering
problem, and bin packing problem are all strongly NP-complete (proofs
by reduction from 3-PARTITION, \citeR<e.g.>{MartelloT90,Fukunaga05}). Thus, these problems cannot be
solved by a polynomial or pseudo-polynomial time algorithm unless
$P=NP$, and the state-of-the-art approach for optimally solving to these problems is
branch-and-bound. In contrast, the single-container 0-1 Knapsack problem and subset sum problem are only 
{\em weakly NP-complete}, and can be solved in pseudopolynomial time using dynamic
programming algorithms \cite{KellererPP04}.

In this section, 
we begin by describing the standard item-oriented branch-and-bound approach to solving
multi-container problems. We then describe bin completion, an alternate, bin-oriented strategy.
For clarity and simplicity of exposition, we detail the algorithms in the context of the bin
packing problem. 

\subsection{Item-Oriented Branch-and-Bound}

\begin{figure}
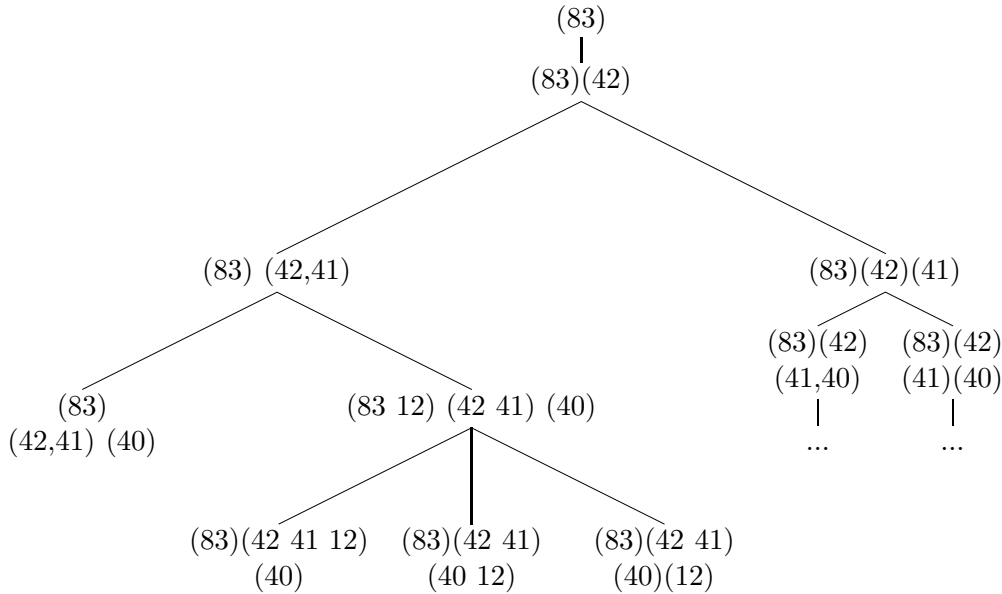

\Tree 
[.(83) 
  [.{(83)(42)} 
    [.{(83) (42,41)}
      {(83)\\(42,41) (40)} 
      [.{(83 12) (42 41) (40)}
        {(83)(42 41 12)\\(40)}
	{(83)(42 41)\\(40 12)}
	{(83)(42 41)\\(40)(12)} ] ]
    [.{(83)(42)(41)} 
      [.{(83)(42)\\(41,40)} {...} ]
      [.{(83)(42)\\(41)(40)} {...} ]
] ] ]

\caption{Partial, item-oriented search space for a bin packing
instance with capacity 100 and items \{83,42,41,40,12,11,5\}. Each
node corresponds to a decision about which existing or new bin an item
is assigned to. Items are considered in nonincreasing order of
weight.}
\label{fig:item-oriented}
\end{figure}

The standard approach to solving multi-container problems is an {\em
item-oriented} branch-and-bound strategy.
%
Suppose we have a bin packing problem instance where the bin capacity
is 100, and we have 7 items with weights 83, 42, 41, 40, 12, 11, and 5.
We perform a branch-and-bound procedure to search the space of
assignments of items to bins, where each node in the branch-and-bound
search tree corresponds to an item, and the branches correspond to
decisions about the bin in which to place the item.  Assuming that we
consider the items in order of non-increasing weight, we first place
the 83 in a bin, resulting in the set of bins $\{(83)\}$.  Next, we
consider the 42. If we put it in the same bin as the 83, resulting in
$\{(83,42)\}$, then we exceed the bin capacity. The other possibility
is to put the 42 in a new bin, resulting in $\{(83),(42)\}$. This is the only feasible choice at this node.
Next, the 41 is assigned to a bin. There are three possible places to
assign the 41: (a) in the same bin as the 83, resulting in
$\{(83,41),(42)\}$, (b) in the same bin as the 42, resulting in
$\{(83),(42,41)\}$, and (c) in a new bin of its own, resulting in
$\{(83),(42),(41)\}$.  Option (a) exceeds bin the capacity of the
first bin and is infeasible. Options (b) and (c) are feasible choices,
so we branch on this item assignment, resulting in two
subproblems, to which the search procedure is recursively applied.
Figure \ref{fig:item-oriented} illustrates a portion of this search space (we show only the feasible nodes).

This branch-and-bound procedure
is {\em item-oriented}
because at each node, we decide upon the placement of a particular
item. Upper and lower bounding techniques specific to the problem can be applied to make this basic depth-first strategy more efficient.
Item-oriented branch-and-bound appears to be the most ``natural'' strategy for multicontainer problems.
The seminal
paper by Eilon and Christofides \citeyear{EilonC71}, which was the first
paper to address optimal solutions for multi-container problems,
proposed an item-oriented branch-and-bound strategy. Most of the
work in the literature on algorithms for optimally solving multi-container
problems have relied upon this item-oriented strategy.

\subsection{Bin Completion, a Bin-Oriented Branch-and-Bound Strategy}
\label{sec:bin-oriented}
An alternate problem space for solving multi-container problems is
{\em bin-oriented}, where nodes correspond to decisions about which
remaining item(s) to assign to the current bin.


A {\em bin assignment} $B = (item_1,...,item_k)$ is a set of {\em all} the items that are assigned to a given bin.
Thus, a valid solution to a bin packing problem instance consists of a set of bin assignments, where each item appears in exactly one bin assignment.
A bin assignment is {\em feasible} with respect to a given bin
capacity $c$ if the sum of its weights does not exceed $c$. Otherwise,
the bin assignment is {\em infeasible}. The definition of
feasibility is the same for the MKP; however, for bin covering and the MCCP, we
define a bin assignment as feasible with respect to a bin quota $q$ if
the sum of its weights is at least $q$. Given a set of $k$ remaining
items, we say that a bin assignment $S$ is {\em maximal} with respect to capacity $c$ if $S$ is
feasible, and adding any of the $k$ remaining items would make it
infeasible. Similarly, for bin covering and the MCCP, a feasible bin
assignment is {\em minimal} with respect to quota $q$ if removing any item would make it
infeasible.
For brevity, in the rest of this paper, we omit the qualification
``with respect to a given capacity/quota'' when it is unnecessary (for
example, in bin packing and bin covering, all bins have the same
capacity/quota so no qualification is necessary).

\begin{figure}
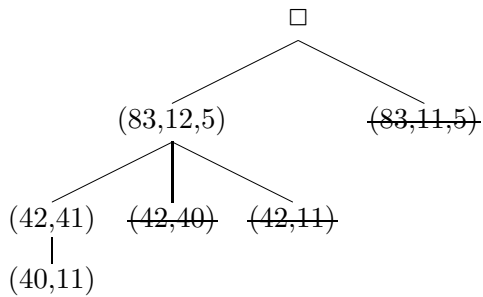

\Tree 
[.$\Box$
  [.(83,12,5)
    [.(42,41)
      (40,11) ] 
    \st{(42,40)}
     \st{(42,11)}  ]
  \st{(83,11,5)} ]
\caption{Bin-completion search space for a bin packing instance with capacity 100 and items \{83,42,41,40,12,11,5\}. Each node represents a maximal, feasible bin assignment for a given bin. Bin assignments shown with a \st{strikethrough}, e.g., (83,11,5), are pruned due to the dominance criterion described in Section \ref{sec:dominance-criteria}.}
\label{fig:bin-completion}
\end{figure}

Bin completion (for bin packing) is a bin-oriented branch-and-bound algorithm in which
each node represents a maximal, feasible bin assignment.
{\em Rather than assigning items one at a time to bins,
bin completion branches on the different maximal, feasible bin
assignments.}
The nodes at a given level in a bin completion search
tree represent different maximal, feasible bin assignments that
include the largest remaining item. 
The nodes at the next level represent different maximal, feasible bin
assignments that include the largest remaining item, etc.  
The reason that we restrict sibling bin assignments in the bin packing search tree to have the largest remaining number in common is to eliminate
symmetries that are introduced by the fact that all bins have the same
capacity in our bin packing fomulation.
Thus, the depth of any branch of the
bin completion search tree corresponds to the number of bins in the
partial solution to that depth.  Figure \ref{fig:bin-completion} shows
an example of a bin completion search tree. 

\subsection{Dominance Criteria for Bin Assignments}
\label{sec:dominance-criteria}

The key to making bin completion efficient is the use of a {\em
dominance criterion} on the feasible bin assignments that requires us to
only consider a small subset of them.

Bin comletion (for bin packing) only considers maximal, feasible assignments, because it is clear
that non-maximal assignments are dominated by maximal assignments,
i.e., assigning a non-maximal assignment to a bin cannot lead to a
solution with fewer bins than assigning a maximal assignment.  For
example, if the current bin has 20 units of remaining space and the
remaining items are $\{15, 30,40,60 \}$, it is {\em always} better to
add the 15 to the current bin, because we must eventually pack the 15
somewhere.  
We now formalize this notion of dominance and describe more powerful
dominance criteria that significantly prune the search space.

\begin{definition}[Dominance]  Given two feasible bin assignments $F_1$ and $F_2$, $F_1$ {\em
dominates} $F_2$ if the value of the optimal solution which can be
obtained by assigning $F_1$ to a bin is no worse than
the value of the optimal solution that can be obtained by assigning $F_2$ to the same bin.
\end{definition}

If we can show that a bin assignment $B$ is dominated by another bin
assignment $A$, then we can prune the search tree under
$B$. Maximality is a trivial dominance criterion for bin
completion. If a feasible bin assignment $B$ is not maximal, then by
definition, it must be a subset of a maximal, feasible subset
$A$, and $B$ is clearly dominated by $A$.
We now consider more powerful dominance criteria.

Suppose we have a bin packing instance with bin capacity 100 and items
\{96,3,4,80,15,12\}.  The sets (96,3) and (96,4) are both maximal, feasible
bin assignments.  If we choose the bin assignment (96,3), the
remaining subproblem has the unassigned items \{80,15,12,4\}. On the other hand, if we choose the bin assignment (96,4), the remaining subproblem is \{80,15,12,3\}.
Clearly, the optimal solution to the subproblem \{80,15,12,4\} must use
at least as many bins as the optimal solution to the subproblem
\{80,15,12,3\}. In other words, the optimal solution in the subtree
under the node (96,4) is at least as good as the optimal solution in
the subtree under (96,3), and therefore there is no need to search
under (96,3) because the bin assignment (96,4) dominates
the bin assignment (96,3).

Christofides, Mingozzi, and Toth 
first proposed a more general form for this dominance criterion 
in the context of the min-cost covering
problem \cite{ChristofidesMT79}. We have reformulated their
criterion in terms of the bin packing problem:

\begin{proposition}[CMT Dominance for Bin Packing] Given two feasible sets $A$ and $B$,  $A$ {\em dominates} $B$ if: (1) $|A| \geq |B|$ and 
(2) there exists a one-to-one (but not necessarily onto) mapping $\pi$ from $B$ to $A$
such that each item $b \in B$, is mapped to an element of $A$ with a weight that is greater than or equal to the weight of $b$, i.e., 
$w(b) \leq w(\pi(b))$.\footnote{Recall that
a function $f: X \rightarrow Y$ is {\em one-to-one} if for any two distinct elements $x, x' \in X$, $f(x) \neq f(x')$.
A function $f:X \rightarrow Y$ is {\em onto} Y if each element of $Y$ is the image under $f$ of some element in $X$. }
\end{proposition}

In other words, if for each element $b$ in a feasible bin assignment $B$, there
is a corresponding item $a$ of a feasible bin assignment $A$ such that $b \leq a$, then $A$ dominates $B$.
The reason is as follows: consider a bin packing solution $S$, where a
single bin is assigned the items in $B$. All of the items
must be assigned to some bin, so the items $a \in A$ are assigned to
some other bin(s).
For each item in $B$, we can swap the corresponding item in $A$, and the resulting solution $S'$ will be feasible, and have no more bins than $S$.
For example, consider a bin packing instance with the items \{10,9,8,7,6\} and bin capacity 20. The bin assignment $A=(10,8)$  dominates the bin assignment $B=(9,7)$, because we can map the 9 to the 10 and map the 7 to the 8.

\citeA{MartelloT90} proposed a more powerful dominance criterion that subsumes the CMT criterion. 
Consider a bin packing instance with items \{6,4,2,1,...\}
and capacity 10. The assignment (6,4) dominates the assignment (6,2,1)
because given any solution with the assignment (6,2,1), we can swap
the 2 and 1 with the 4, resulting in a solution with the assignment
(6,4) and the same number of bins. The CMT criterion does not account
for this.
More generally, consider two feasible bin assignments $A$ and $B$.
If all the elements of $B$ can be
packed into bins whose capacities are the elements of $A$, then set
$A$ dominates set $B$. For example, let $A = (20,30,40)$ and let
$B=(5,10,10,15,15,25)$. Partition $B$ into the subsets
$(5,10)$,$(25)$, and $(10,15,15)$. Since 5 + 10 $\leq$ 20, $25 \leq
30$, and $10 + 15 + 15 \leq 40$, set $A$ dominates set $B$. 
More formally:

\begin{proposition}[Martello-Toth Bin Packing Dominance Criterion] 
Let $A$ and $B$ be two feasible bin assignments. $A$ {\em dominates}
$B$ if $B$ can be partitioned into $i$ subsets $B_1,...B_i$ such that
each subset $B_k$ is mapped one-to-one to (but not necessarily onto)
an item $a_k \in A$
such that the sum of the weights of the items in
$B_k$ is less than or equal to the weight of $a_k$.
\end{proposition}

\noindent Proof: Suppose we have candidate solution $s$ 
that assigns $B$ to bin $m$, and all of the bin assignments in $s$ are currently feasible.
Let $A$ be a feasible bin assignment such that $B$ can be partitioned into $i$ subsets
$B_1,...,B_i$, and each subset $B_k$ can be mapped one-to-one to $a_k \in A$, 
such that the sum of the weights of the items in $B_k$ is less than or equal to the weight of $a_k$.
Consider swapping $B$ with $A$.
Each subset $B_k$ is swapped with its corresponding element $a_k$ (where $a_k$ is assigned to bin $d_k$ in the original solution $s$).
Since $A$ is feasible, bin $m$ remains feasible after $A$ and $B$ are swapped.
Now consider each of the bins $d_1,...d_i$ where the subsets $B_1,...,B_i$ end up after the swaps (in other words, $d_1,...d_i$ are the bins which contain $a_1,...,a_k$, respectively, in the original solution $s$). Each such target bin $d_k$ will remain feasible after the swap, since (1) the bin assignment of $d_k$ is
feasible prior to the swap, and (2) the sum of the weights of the items in
$B_k$ is less than or equal to the weight of $a_k$.
Thus, the resulting solution $s'$ after the swap is 
(a) feasible (all bin assignments are feasible) (b) 
no worse than the initial solution $s$,  and (c) assigns $A$ to bin $m$.  Therefore,  $A$ dominates $B$.
$\Box$



The Martello-Toth dominance criterion is a generalization of the CMT
dominance criterion -- the CMT dominance criterion is a
special case of the Martello-Toth criterion where we only consider
partitions of $B$ into single-element subsets.  Thus, any node
that would be pruned by the CMT criterion is also pruned by the Martello-Toth criterion, but not vice versa.


Similarly, we can define a dominance criterion for bin covering as follows: 

For bin covering (and the MCCP), a 
bin assignment is feasible with respect to a given bin if the sum of the item weights is greater than or equal to the bin quota $q$.

\begin{proposition}[Bin Covering Dominance Criterion]
\label{def:bc-dominance-criterion}
Let $A$ and $B$ be two feasible
assignments.  $A$ dominates $B$ if 
$B$ can be partitioned into $i$ subsets $B_1,...,B_i$
such that each item $a_k \in A$ is mapped one-to-one to
(but not necessarily onto) 
a subset $B_k$, such that the weight of each $a_k$ is less than or equal to the sum of the item weights of the corresponding subset $B_k$ (i.e., $B_k$ ``covers'' $a_k$).
\end{proposition}

\noindent Proof: Suppose we have a solution $s$ 
that assigns $B$ to bin $m$, and all of the bin assignments in $s$ are currently feasible.
Let $A$ be a feasible bin assignment such that $B$ can be partitioned into $i$ subsets
$B_1,...,B_i$, and each item $a_k \in A$ can be mapped one-to-one to a subset $B_k$, 
such that the weight of $a_k$ is less than or equal to the sum of the weights of the items in $B_k$.
Consider swapping $B$ with $A$.
Each subset $B_k$ is swapped with its corresponding element $a_k$ (where $a_k$ is assigned to bin $d_k$ in the original solution $s$).
Since $A$ is feasible, bin $m$ remains feasible after $A$ and $B$ are swapped.
Now consider each of the bins $d_1,...,d_{|A|}$ where the subsets $B_k$ end up after the swaps. In other words, $d_1,...,d_{|A|}$ are the bins which contain $a_1,...,a_k$, respectively, in the original solution $s$. Each such target bin $d_k$ will remain feasible after the swap, since (1) the bin assignment of $d_k$ is
feasible prior to the swap, and (2) the sum of the weights of the items in
$B_k$ is greater than or equal to the weight of $a_k$ (and thus the quota of bin $d_k$ will continue to be satisfied).
Thus, the resulting solution $s'$ after the swap is 
(a) feasible (all bin assignments are feasible) (b) 
no worse than the initial solution $s$,  and (c) assigns $A$ to bin $m$.  Therefore,  $A$ dominates $B$.
$\Box$


The dominance criteria for the MKP and MCCP are similar to the
dominance criteria for bin packing and bin covering, respectively,
except that we must also take into consideration the
profits/costs. The proofs are similar to the proofs for bin packing
and bin covering.

\begin{proposition}[MKP Dominance Criterion]
\label{def:mkp-dominance-criterion}
Let $A$ and $B$ be two assignments that are feasible with respect to capacity $c$.
$A$ {\em dominates} $B$ if $B$
can be partitioned into $i$ subsets $B_1,...,B_i$ such that each
subset $B_k$ is mapped one-to-one to
(but not necessarily onto) 
$a_k$,
an element of $A$, and for all $k \leq i$, (1) the weight of $a_k$ is
greater than or equal the sum of the item weights of the items in
$B_k$, and (2) the profit of item $a_k$ is greater than or equal to the
sum of the profits of the items in $B_k$.
\end{proposition}

\begin{proposition}[MCCP Dominance Criterion]
\label{def:mccp-dominance-criterion}
Let $A$ and $B$ be two assignments that are feasible with respect to quota $q$.
$A$ dominates
$B$ if $B$ can be partitioned into $i$ subsets
$B_1,..., B_i$ such that 
each item $a_k \in A$ is mapped one-to-one 
(but not necessarily onto) 
to a subset $B_k$, and for each $a_k \in A$ and its corresponding subset $B_k$,
(1) the weight of $a_k$ is less than or equal to the sum of the
item weights of the items in $B_k$, and (2) the cost of item $a_k$
is less than or equal to the sum of the cost of the items in $B_k$.
\end{proposition}

The CMT dominance criterion for the MCCP originally proposed in \cite{ChristofidesMT79} is a special case of Proposition \ref{def:mccp-dominance-criterion}, 
where $|B_k|=1$ for all $k$.

Note that in the ``packing'' problems such as the MKP and bin packing, the
dominance criteria require that
the subsets of the dominated assignment are packed into the items of the dominating assignment.
In contrast, with the ``covering'' problems such as the MCCP and bin covering,
the dominance criteria requires that the subsets of the dominated assignment ($B$) cover the items in the dominating assignment ($A$).


\subsubsection{Bin-Oriented Branch-and-Bound + Dominance = Bin Completion}

At this point, we have defined the salient features of the bin completion 
approach to solving multicontainer packing and knapsack problems:
\begin{itemize}
\item A bin-oriented branch-and-bound search where the nodes correspond
to maximal (or minimal), feasible bin assignments; and
\item The exploitation of a dominance criterion among bin assignments to prune the search space.
\end{itemize}

The first instance of a bin completion algorithm that we are aware of is
the Christofides, Mingozzi, and Toth algorithm for the min-cost covering problem in 1979
\cite{ChristofidesMT79}, which used the CMT criterion described above.
However, as far as we know, no further research was done with bin
completion algorithms until our work on bin completion for bin
packing \cite{Korf02}.

The Martello-Toth dominance criterion was proposed by Martello and Toth
\citeyear{MartelloT90}, as a component of the Martello-Toth Procedure (MTP), a
branch-and-bound algorithm for bin packing.
However, the MTP
branch-and-bound algorithm is item-oriented, and they only exploit
this dominance property in a limited way. In particular, they take
each remaining element $x$, starting with the largest element, and check if
there is a {\em single} assignment of $x$ with one or two more elements that
dominates all feasible sets containing $x$. If so, they place $x$ with
those elements in the same bin, and apply the reduction to the
remaining subproblem. They also use dominance relations to prune some
element placements as well. 

Another earlier instance of a bin completion algorithm 
is the BISON algorithm for bin packing
by Scholl, Klein, and J\"{u}rgens \citeyear{SchollKJ97}.  
BISON uses the following, very limited form of the Martello-Toth dominance criterion: if a bin assignment has one or more items that can be replaced by a single free item without decreasing the sum, then this assignment is dominated.
It is interesting that despite the fact that the basic idea of bin-oriented search with dominance-based pruning was demonstrated by Christofides, Mingozzi, and Toth, both the MTP and BISON only use a limited
form of the Martello-Toth dominance criterion, and the two ideas were not successfully integrated until our work.\footnote{As shown in \cite{Korf02,Korf03}, bin-oriented search using the full Martello-Toth dominance criterion results in dramatic speedups compared to the MTP.} Presumably, the reason is that it is not trivial to generate undominated bin assignments efficiently.

\subsection{Generating Undominated Bin Assignments}
\label{sec:generating-assignments}

A key component in bin completion is the efficient generation of
undominated bin assignments. 
%
An obvious approach is to generate all feasible bin assignments, then apply the dominance tests to eliminate the dominated assignments. However, this is impractical, because the number of feasible assignments is exponential in the number of remaining items, and the memory and time required to generate and store all assignments would be prohibitive.
We now describe an algorithm that generates {\em all and only undominated bin assignments}, which enables the efficient implementation of bin completion.


We can generate all subsets of $n$ 
elements by recursively traversing a binary tree.  
Each internal node corresponds to an item, 
where the
left branch corresponds to the subsets that include the item, and the
right branch corresponds to the subsets that do not include the item.
Thus, each leaf node represents an individual subset.
Note that this binary tree is the search space for this subproblem (generating undominated bin assignments), and is distinct
from the search space of the higher level bin-completion algorithm
(i.e.. the space of undominated maximal assignments).

Given $n$ elements and a container capacity $c$, the {\em feasible bin
assignments} are the subsets of the $n$ elements,  the sum of whose weights
do not exceed $c$.  The recursive traversal described above for
generating all subsets can be modified to generate only feasible
assignments as follows: At each node in the tree, we keep track of the
sum of items that we have committed to including in the subset (i.e.,
the sum of the weights of the items for which we have taken the left
branch).  During the recursive traversal of the tree, we pass a
parameter representing the remaining capacity of the bin.
Each time a left
branch is taken (thereby including the item), we reduce the remaining capacity by
the item weight.  If the remaining capacity drops to zero or less, we prune the tree
below the current node, since we have equaled or exceeded the container capacity.
Thus, this algorithm
generates only feasible bin assignments.


Now, we further extend the algorithm to 
generate only the
{\em undominated}, feasible bin assignments.  Suppose that we have a
feasible set $A$ whose sum of weights is $t$.  
The {\em excluded items} are all items that are not in $A$.
Set $A$ will be dominated (with respect to the Martello-Toth dominance criterion) if and only if it contains any subset whose
sum $s$ is less than or equal to an excluded item $x$, such that
replacing the subset with $x$ will not exceed the bin capacity $c$.
This will be the case if and only if there exists an excluded item $x$ and a subset with weight sum $s$ such that $t-s+x \leq c$. 
Therefore, to check if $A$ is undominated, we enumerate each possible subset of its items, and for each subset,
we compare it against each excluded number $x$ to verify that $t-s+x > c$, where $s$ is the sum of the item weights of the subset. If so,
then $A$ is undominated, and we store $A$ in a list of undominated assignments; otherwise, $A$ is dominated. 
Further optimizations can be found in \cite{Korf03}. 

This algorithm generates feasible bin assignments and immediately tests them for
dominance, so it never stores multiple dominated bin assignments. Furthermore,
the dominance test is done by comparing the included elements to
excluded elements, and {\em does not involve any comparison between a
candidate bin assignment and previously generated bin
assignments}. Therefore, the memory
required for dominance testing is linear in the number of items. 
In contrast, a method that depends on comparisons between candidate sets,
such as the earlier algorithm described in \cite{Korf02}, requires
memory that is linear in the number of undominated bin assignments,
which is potentially exponential in the number of items.
The ability to incrementally
generate undominated bin assignments using linear space without having to store all of the undominated assignments enables the
hybrid incremental branching strategy, described in Section
\ref{sec:hybrid-incremental}.

\section{Extensions to Bin Completion}
\label{sec:extensions}

We now describe some extensions to bin completion that significantly
improve search efficiency. 
Again, for clarity, we describe the
algorithms mostly in the context of bin packing, but as with the basic bin completion algorithm, these extensions can be adapted
straightforwardly to the multiple knapsack, bin covering, and min-cost
covering problems.

\subsection{Nogood Pruning (NP)}
\label{sec:NP}

Nogood pruning prunes redundant nodes in the bin
completion search tree by detecting symmetries. 
Since we need to refer to specific bins, we extend our notation for
bin assignment. Let $(A)_d$ denote a bin at depth $d$ that is
assigned the elements in  $A$. Thus, $(10,8,2)_1$ and
$(10,7,3)_1$ denote two possible bin assignments for the bin at depth 1.

Suppose we have an instance with the numbers
\{10,9,8,7,7,3,3,2,2\},
and bin capacity c=20.  
After exhausting the subproblem below the assignment
$(10,8,2)_1$, and while exploring the subproblem below the
assignment $(10,7,3)_1$, assume we find a solution that assigns
$(9,8,2)_2$. We can swap the pair of items (8,2) from the assignment $(9,8,2)_2$ with the
pair of items (7,3) from the assignment $(10,7,3)_1$, resulting in a solution with
$(10,8,2)_1$ and $(9,7,3)_2$ and the same number of bins. 
However, we have already
exhausted the subtree below $(10,8,2)_1$ and we would have found a solution with the same number of bins as the best solution in the subtree below $(9,7,3)_2$.
Therefore, we can
prune the branch below $(9,8,2)_2$, because it is redundant (in other words, we have detected that the current partial solution is symmetric to a partial state that has already been exhaustively searched).

More formally, let
$\{N_1,N_2,...,N_m\}$ be a set of sibling nodes in the search tree, and
let $\{S_1,S_2,...,S_m\}$ be the bin assignments for 
each sibling node, excluding the first item assigned to the bin, which is common to all the sibling nodes.
When searching the subtree below node $N_i$ for $i > 1$, we exclude
any bin assignment $B$ that (1) includes all the items in $S_j$, and
(2) swapping the items in $S_j$ from $B$ with the items $S_i$ in $N_i$
results in two feasible bin assignments, for $i>j$.  The items in
$S_j$ become a {\em nogood} with respect to nodes deeper in the
tree.\footnote{While the term {\em nogood} in the constraint
programming literature often refers to an assignment of value(s) to
variable(s) that cannot lead to a solution, we use the term to mean
an assignment that cannot lead to any solution that is better than a
previously discovered solution, similar to the usage in
\cite{FocacciS02}.}

If there exists such a bin assignment $B$, then we could swap the
items $S_j$ from $B$ with the items $S_i$ in $N_i$, resulting in a
partial solution with the bin assignment $S_i$ in bin $N_i$. However,
we have already exhausted the subtree below $N_i$, so this is a
redundant node and can be pruned.


As the search progresses down the tree, a list of nogoods is
maintained, which is the set of bin assignments against which each
candidate undominated bin assignment is compared.  
Given a candidate bin assignment $B$, where the items are sorted
according to weight, the current implementation of the test for nogood
pruning compares the items against each nogood. Since the items in the
nogood sets are also sorted by weight, each comparison takes time
linear in the cardinality of $B$. In the worst case, the number of
nogoods that we must compare with a candidate assignment at level $d$
corresponds to the number of undominated bin assignments at levels
$1,...,d-1$ that are currently on the stack (that is, all of the ancestors and the siblings of the ancestors of the current node).
Note that the list of nogoods need not grow monotonically as we go down the search tree.  If at any point, a nogood $N$ is no longer a subset of the set of remaining items, then $N$ has been 'split', and  can be removed from the nogood list that is passed down the search tree.

\subsection{Nogood Dominance Pruning (NDP)}
\label{sec:NDP}

The following {\em nogood dominance pruning} (NDP) technique allows even more pruning:
Suppose that after exhausting the subproblem below the assignment $(10,8,2)_1$, and
while exploring the subproblem below the assignment $(10,7,3)_1$, 
we consider the assignment $(9,7,2)_2$.
We can swap the pair of items (7,2) from bin 2 with the pair of items (7,3) from bin 1 and end up with a solution
with $(10,7,2)_1$ and $(9,7,3)_2$.
However, according to the Martello-Toth dominance criterion, 
(10,7,2) is dominated by (10,8,2), and we have 
already exhausted the search below the node 
$(10,8,2)_1$, so we can prune the search under $(9,7,2)_2$ because it is not possible to
improve upon the best solution under $(10,8,2)_1$.

In general, 
given a node with more than one child, when searching the
subtree of any child but the first, we don't need to consider
assignments that are dominated by a bin assignment in a
previously explored child node.  More precisely, let
$\{N_1,N_2,...,N_m\}$ be a set of sibling nodes in the search tree, and
let $\{S_1,S_2,...,S_m\}$ be the corresponding sets of items 
used in the bin assignment at each node.
When searching the subtree below node
$N_i$ for $i > 1$, we exclude any bin assignment $A$ for which there exists an assignment $S_j$, $j<i$, such that (1) $A$ is dominated
by the items in $S_j$ (note that an assignment dominates
itself), and (2) $A$ can be swapped with $S_i$, such that the resulting bin assignments are both feasible.  
If there exists such a bin assignment $A$, then we could swap the
items from $A$ with the items $S_i$ in $N_i$, resulting in a
partial solution with the bin assignment $A$ in bin $N_i$, and the same number of bins. However,
since $A$ is dominated by $S_j$, it means that we are searching a node
that is symmetric to one that is dominated by $S_j$, and therefore, it
is not possible to find a solution better than the best solution under
$S_j$, so $A$ can be pruned.



We can also describe nogood dominance pruning in terms
of a more general constraint programming formulation, where variables
correspond to items and values denote the bins to which they
are assigned.
Given a
partial, $j$-variable solution $x$, nogood dominance pruning tries to
show, via swapping of items and dominance checks, that $x$ is in the same equivalence class as another partial
solution $x'$ such that $\bar{x}'_{i}$, the subset of $x'$ including
the first $i$ variables, is dominated by another partial solution $q$.
Thus, if we have exhausted the subtree below $q$ in the search
tree, we do not need to search the subtree below $x'$.


Nogood dominance pruning is strictly more powerful than 
nogood
pruning.  Any node pruned by nogood pruning will be pruned by nogood dominance pruning, but
not vice versa.  Of course, since NDP must detect dominance
relationships as opposed to equivalence relationships, NDP will incur
more overhead per node compared to NP.  Our current
implementation propagates a list of nogood sets along the tree. While
generating the undominated completions for a given bin, we check each
one to see if it is dominated by any current nogood. If so, we ignore that bin assignment. 
Our current implementation uses a brute-force algorithm to check for dominance between a candidate bin assignment and each nogood, which in the worst case takes time exponential in the cardinality of the bin assignment (for each nogood).

Since nogood pruning is much less expensive than nogood dominance
pruning, we use a combined pruning strategy. Whenever we apply NDP, we
actually apply both NP and NDP at each node. First, the candidate bin
assignment is checked against a nogood, and nogood pruning is
applied. We apply nogood dominance pruning only if the bin assignment was not pruned by nogood pruning. {\em Thus, we never pay the
full cost of applying NDP to nodes that can be pruned quickly by NP}.
As with nogood pruning, NDP requires storing a set of nogoods, and the number of possible nogoods at a particular search depth $d$ in the worst case is the number of undominated bin assignments considered at depths $1,...,d-1$.
Note that when using NDP, we do not apply the optimization described
above in Section \ref{sec:NP} for removing nogoods from the nogood list that are passed down
the tree when using nogood pruning. The reason is that even if a
nogood has been ``split'' and can no longer prune a bin assignment due
to nogood pruning, that nogood may still be able to prune a bin assignment
due to nogood dominance pruning.

The size of the nogood list increases with depth, and we compare
each bin assignment against each nogood. Therefore, the per-node overhead of NDP
increases with depth.  This means that pruning at the bottom of the
tree (where pruning has the lowest utility) is more expensive than
pruning at the top of the tree (where pruning has the highest
utility).  A simple strategy which address this issue is {\em
depth-limited NDP}, where NDP is applied only at nodes up to
the NDP depth limit $L$. At nodes below the depth limit, 
only the weaker nogood pruning is applied. In the experiments described in this paper, we did not use this depth-limited strategy because NDP consistently outperformed nogood pruning without having to use depth limits.


\subsection{Related Work in Constraint Programming}
\label{sec:related-work-in-cp}

Nogood pruning identifies and prunes nodes by
detecting whether the bin assignment for the current node contains a
nogood. This is related to other symmetry-breaking techniques
proposed in the constraint programming literature.
A {\em symmetry} partitions the set of possible assignments of values to
variables into equivalence classes \cite{GentS00}.  The goal of symmetry
breaking is to prune nodes that can be mapped to a previously explored
node via a symmetry function. 
Symmetry-breaking approaches introduce constraints during the search that prune
symmetric variable assignments \cite<e.g.,>{GentS00,FahleSS01,FocacciM01}.
Similarly, our nogood and nogood dominance pruning techniques
dynamically introduce constraints that prune variable assignments that cannot
lead to a solution that is better than the best solution found so far.
Nogood dominance pruning uses the same dynamic nogood recording
mechanism as nogood pruning, but goes a step further in detecting
dominance relationships based on the nogoods.  The general notion of dominance exploited by NDP is more powerful than symmetry, since dominance can be asymmetric, i.e., all symmetries are dominance relationships, but not vice versa.

Our NDP technique is similar to the pruning
technique proposed by Focacci and Shaw \citeyear{FocacciS02} for
constraint programming, who applied their technique to the symmetric
and asymmetric traveling salesperson problem with time windows.  
Both methods attempt to prune the search by proving
that the current node at depth $j$, which represents a partial
$j$-variable (container) solution $x$, is dominated by some previously explored
$i$-variable partial solution (nogood), $q$.
The main difference between the two methods is the approach used to
test for dominance. Focacci and Shaw's method extends $q$ to a
$j$-variable partial solution $q'$ which dominates $x$.  They apply a
local search procedure to find the extension $q'$.

In contrast,
our NDP method starts with a partial, $j$-variable solution $x$ and tries to
transform it to a partial solution $x'$ such that $\bar{x}'_{i}$, the subset
of $x'$ including the first $i$ variables, is dominated by $q$.  We do
this by swapping the values of the $i$th and $j$th variables in $x$ to
derive $x'$, and testing whether $\bar{x}'_{i}$ is dominated by $q$.

For 
efficiency, the current implementations
of both nogood dominance pruning methods are {\em weak}, in the sense
that if $x$ is dominated by $q$, the procedures will not necessarily
detect the dominance. 
Focacci and Shaw rely on an incomplete, local search to find the 
extension $q'$.
Due to the cost of considering the transformations, we only 
consider transformations involving two variables (containers), but to fully exploit the dominance criterion, we
would need to consider transformations involving all variables  $i,i+1,...,j$.

\subsection{ Reducing the Computation Required Per Node}
\label{sec:hybrid-incremental}
An issue with enumerating all undominated completions
and applying value ordering is that computing the
undominated sets is 
itself NP-complete.
For a multicontainer problem instance where the items have weight $w_i$ and average container capacity is $c$, the average number of items that fit in a container is  $x = \sum_{i=0}^{n} w_i / c$.
The time to generate all undominated bin assignments
increases with $x$.
This is not an issue for 
bin packing, where problems with large $x$ tend to be
easily solved using heuristics such as best-fit decreasing. The
solution found by the heuristic often equals the lower bound. Therefore,
such instances require no search, and hence there is no need to compute undominated bin assignments.
On the other hand, for multiple knapsack and bin covering,
we have observed experimentally that it is much less
likely that heuristics will match the optimistic bound and allow
termination without search, and we have found that for instances with high $x$,
the algorithm would not terminate within a reasonable time limit because it
was spending an inordinate amount of time computing the set of
undominated completions at each node. In addition, generating all undominated bin assignments may cause us to run out of memory. 

An alternative approach is to start to go down the search tree and
explore some of the children of a node without first enumerating and sorting
all the children. In cases where (a) a good optimistic bound is
available, and (b) it takes relatively little search to find an optimal solution, 
this approach leads to dramatic speedups compared to the
original scheme of generating all undominated completions before going further
down the search tree. The price we pay for this strategy is that we lose the benefits of
ordering all the candidate bin assignments according to the value-ordering
heuristic. 

To solve this problem, we propose a {\em hybrid incremental branching
strategy} that generates $h$ children at each node, applies the
value-ordering heuristic to these, then recursively calls bin
completion on the remaining subproblems.  For each node, we first
generate $h$ children, and sort them according to the value-ordering
heuristic. Then, we explore the subtrees under each of these
children. After the subtrees under the first $h$ children are fully
explored, then the next $h$ children are generated and their subtrees
are explored, and so on.

\begin{figure}
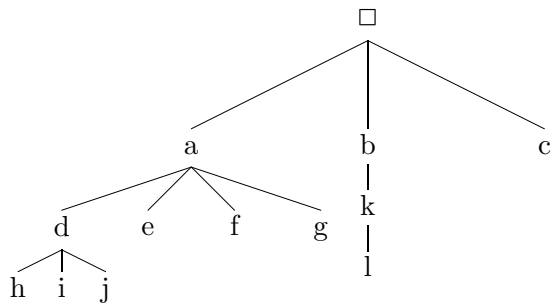

\Tree
[.$\Box$
  [.a
     [.d 
        h i j ]
     e f g ]
  [.b
     [.k l ] ]
  c ]
\caption {Hybrid Incremental Branching Strategy}
\label{fig:hybrid-incremental}
\end{figure}

For example, consider the tree in Figure \ref{fig:hybrid-incremental}, where the nodes correspond to undominated bin assignments. Assume that this is the complete search tree, and assume for simplicity that there is no value ordering heuristic.
The standard bin completion algorithm first generates the children of the root:  $a$, $b$, and $c$. 
Then, bin completion selects one of the children (say $a$), and expands $a$'s children ($d$,$e$,$f$,$g$).
It then generates the children of $d$, and so on. Thus, the order of node generation is: $a, b, c, d, e, f, g,h, i, j, k, l$. This is a {\em pre-order traversal} of the tree.
Now consider hybrid incremental branching with width 1. This corresponds to a standard postorder traversal of the tree, where the order in which nodes are generated is $a, d, h, i, j,  e, f,  g, b, k, l, c$.
Hybrid incremental branching with width 2 generates the nodes in the order $a, b, d, e, h, i, j, f, g, k, l, c$.

\section{The Multiple Knapsack Problem (MKP)}
\label{sec:multiple-knapsack}

Given a set of containers and a set of items (characterized by a
weight and profit), the objective of the multiple knapsack problem (defined formally in section \ref{sec:mkp-definition}) is
to assign items to containers such that the container capacities are
not exceeded, and the sum of the profits of the items assigned to the
containers is maximized.



We compared bin completion to two standard algorithms: the state-of-the-art Mulknap algorithm \cite{Pisinger99}, as well as the MTM algorithm \cite{MartelloT81}.

\subsection{The MTM Algorithm}
\label{sec:mtm}
The MTM algorithm of Martello and Toth \citeyear{MartelloT81}
is an item-oriented
branch-and-bound algorithm.
The items are ordered
according to non-increasing {\em efficiency} (ratio of profit to weight), so that the next item selected by the variable-ordering heuristic for the item-oriented branch-and-bound 
is the
item with highest efficiency that was assigned to at least one
container by a greedy bound-and-bound procedure (see below).  The branches
assign the selected item to each of the containers, in order of
non-decreasing remaining capacity.

At each node, an upper bound is computed using a relaxation of the MKP, which 
is obtained by 
combining all of the remaining $m$ containers in the MKP into a single container with aggregate capacity $C = \sum_{i=1}^{m}c_i$,
resulting in the single-container, 0-1 knapsack problem:

\begin{align}
\mbox{maximize} &\sum_{j=1}^{n} p_j x'_j \\
\mbox{subject to} 
&\sum_{j=1}^{n} w_jx'_j \leq C, \\
& x'_j \in \{0,1\}, j=1,...,n.
\end{align}

\noindent where the variable $x'_j$ represents whether item $j$ has been assigned
to the aggregated bin.  This {\em surrogate relaxed MKP} (SMKP) can be
solved by applying any algorithm for optimally solving the 0-1 Knapsack problem, and the
optimal value of the SMKP is an upper
bound for the original MKP.  
Thus, this upper bound computation is
itself solving an embedded, weakly NP-complete (single-container) 0-1 Knapsack problem
instance as a subproblem. 
The SMKP is currently the most effective
upper bound for the MKP (details on how the formulation above is derived from an initial surrogate relaxation are in  \citeR{MartelloT81,KellererPP04}).

At each node, the MTM algorithm applies an extension to branch-and-bound called {\em bound-and-bound}.
Consider a branch-and-bound procedure for a maximization problem. If the upper (optimistic) bound computed at a node is less than or
equal to the best known lower bound for the problem, then the node can be pruned. In the standard branch-and-bound approach, the lower bound is simply the objective function value of the best solution found so far. 

In bound-and-bound, we attempt to {\em validate} the upper bound by
applying a fast, heuristic algorithm in order to find a solution whose score equals the upper bound. If such a
solution can be found, then it means that we have found the optimal
solution under the current node (i.e., the upper bound has been validated), and we can backtrack.
On the other hand, if no such solution is found, then we must continue to search under the node.
The MTM algorithm applies a greedy heuristic algorithm for the MKP, which 
involves solving a series of $m$ 0-1 Knapsack problems.  First,
container $i=1$ is filled optimally using any of the remaining items,
and the items used to fill container $i=1$ are removed.  Then,
container $i=2$ is filled using the remaining items. This process is
iterated $m$ times, at which point all containers are filled.

\subsection{The Mulknap Algorithm}
\label{sec:mulknap}
The previous state-of-the-art algorithm for the MKP is 
Mulknap \cite{Pisinger99}.  Like MTM, Mulknap is an item-oriented
branch-and-bound algorithm using the SMKP upper bound and
bound-and-bound.  Mulknap differs from MTM in that it (1) uses a
different validation strategy for the bound-and-bound based on
splitting the SMKP solution, (2) applies item reduction at
each node, and (3) applies capacity tightening at each node. 

Like MTM, Mulknap uses a bound-and-bound strategy, but it uses a different 
approach to validating the upper bound: As described above, the upper
bound for a MKP instance can be computed by solving the surrogate
relaxation of the MKP, or SMKP, which is a 0-1 Knapsack instance.
Suppose we have just computed the optimal solution to the SMKP. Now,
suppose we are able to partition the items used in the SMKP into the
$m$ containers in the original MKP, such that each item used in
the SMKP solution is assigned to some container, and the capacity
constraints are not violated. In this case, we have a solution
to the original MKP that achieves the upper bound.
Details on the capacity tightening and reduction procedures are given by \citeA{Pisinger99}.

\subsection{Bin Completion Algorithm for the MKP}

We now describe our bin completion algorithm for the MKP. 
We apply a depth-first, bin completion branch-and-bound
algorithm.  
Each node in the search tree represents a maximal, feasible bin
assignment for a particular bin. Note that in the MKP, bin capacities can
vary, so at each node, we consider all undominated bin assignments for
the bin. This is in contrast to bin packing \ref{sec:bin-oriented},
where (assuming all bins have identical capacity), we can restrict all
nodes at the same level to have the same (largest) item.

At each node, an upper bound for the remaining
subproblem is computed using the surrogate relaxed MKP (SMKP) bound. The SMKP bound is computed using a straightforward branch-and-bound algorithm with the Martello-Toth U2 upper bound \cite{MartelloT77}.
 Pisinger's R2
reduction procedure \cite{Pisinger99} is applied at each node in order to possibly
reduce the remaining problem.  Then, we select the bin with smallest remaining
capacity (i.e., we use a smallest-bin-first variable ordering heuristic, with ties broken randomly), and its undominated bin assignments are computed and explored
using our dominance criterion (Proposition \ref{def:mkp-dominance-criterion}).
Nogood pruning and nogood dominance
pruning are applied as described in Sections \ref{sec:NP} and \ref{sec:NDP}.





The order in which we branch on the undominated children of the current node has a significant impact on the algorithm's performance. At each node, all of the undominated children of the node are generated and ordered using a {\em value ordering heuristic}. 
We evaluated 11 different value ordering heuristics
and found that the best performing heuristic overall was the  {\em min-cardinality-max-profit} ordering, where 
candidate bin assignments are sorted in order of non-decreasing cardinality and ties are broken according to non-increasing order of profit. Fukunaga \citeyear{Fukunaga05} provides further details on experiments evaluating various value and variable ordering strategies.
Figure \ref{fig:mkp-pseudocode} shows an outline of our bin completion algorithm for the MKP.

\begin{figure}
\begin{center}
\fbox{\begin{minipage}{\pseudocodewidth}{
\begin{tt}
\begin{small}
\begin{tabbing}
m\=m\=m\=m\=m\=m\kill

{\bf MKP\_bin\_completion}(bins,items)\\
\>bestProfit = $-\infty$\\
\>{\bf search\_MKP(bins,items,0)}\\
\\

{\bf search\_MKP}(bins, items,sumProfit) \\

\>if bins==$\emptyset$ or  items == $\emptyset$ \\
\>\>/*we have a candidate solution*/\\
\>\>if sumProfit > bestProfit \\
\>\>\>bestProfit = sumProfit\\
\>\>return\\
\\
\>/* Attempt to reduce problem. reducedBinAssignments are \\
\>\>\> maximal, feasible assignments of items to bins */\\
\>reducedBinAssignments  = {\bf reduce}(bins,items) \\
\>ri = {\bf get\_items}(reducedBinAssignments) /* items eliminated by reduction */\\
\>rb = {\bf get\_bins}(reducedBinAssignments) /* bins eliminated by reduction */\\
\>if reducedItems $\neq$ $\emptyset$\\
\>\>{\bf search\_MKP}(bins $\setminus$ rb, items $\setminus$ ri, sumProfit+$\sum_{i  \in ri}$ {\bf profit}(i)) \\
\>\>return\\
\\
\>/* Attempt to prune based on upper bound */\\
\>if (sumProfit + {\bf compute\_upper\_bound}(items,bins)) < bestProfit \\
\>\>return\\
\\
\>bin = {\bf choose\_bin}(bins)\\
\>undominatedBinAssignments = {\bf generate\_undominated}(items,capacity(bin))\\
\>foreach A $\in$ {\bf sort\_assignments}(undominatedBinAssignments) \\
\>\>if not({\bf nogood}(A)) and not({\bf nogood\_dominated}(A))\\
\>\>\>assign undominated\_bin\_assignment to bin\\
\>\>\>{\bf search\_MKP}(bins $\setminus$ bin, items $\setminus$ items\_in(A),sumProfit+$\sum_{i \in A}${\bf profit}(i))\\

\end{tabbing}
\end{small}
\end{tt}
}\end{minipage}}
\end{center}
\caption{Outline of bin completion for the multiple knapsack problem.
{\tt compute\_lower\_bound} returns the SMKP upper bound on profit for the remaining subproblem, and {\tt reduce} applies Pisinger's R2 reduction to eliminate items and bins if possible.
{\tt choose\_bin} selects the bin with least remaining capacity, and and {\tt sort\_assignments} sorts the undominated bin assignments in order of non-decreasing cardinality, and ties are broken in order of non-increasing profit. 
{\tt generate\_undominated} generates undominated bin assignments using the
algorithm in (Section \ref{sec:generating-assignments}).
{\tt nogood} and {\tt nogood\_dominated} apply nogood pruning (Section \ref{sec:NP}) and
nogood dominance pruning (Section \ref{sec:NDP}).  
}
\label{fig:mkp-pseudocode}
\end{figure}

\subsection{Experimental Results}
\label{sec:mkp-results}

We evaluated our MKP algorithm using the same four classes of instances used by Pisinger \citeyear{Pisinger99}.
We considered:
\begin{itemize}
\item {\em uncorrelated instances}, where the profits $p_j$ and weights $w_j$ are uniformly
distributed in $[min,max]$.
\item {\em weakly correlated instances}, where the $w_j$ are uniformly distributed in
[min,max] and the $p_j$ are randomly distributed in [$w_j - (max-min)/10, w_j + (max-min)/10$]
such that $p_j \geq 1$,
\item {\em strongly correlated instances}, where the $w_j$ are uniformly distributed in
[min,max] and $p_j = w_j + (max-min)/10$,
 and
\item {\em multiple subset-sum instances}, where the $w_j$ are uniformly distributed in
$[min,max]$ and $p_j = w_j$.
\end{itemize}

%
The bin capacities were set as follows:
The first $m-1$ capacities $c_i$ were uniformly distributed in 
$ [ 0.4 \sum_{j=1}^n w_j/m, 0.6 \sum_{j=1}^n w_j/m ]$ for $i = 1,...,m-1$.
The last capacity $c_m$ is chosen as 
$c_m = 0.5 \sum_{j=1}^{n} w_j - \sum_{i=1}^{m-1} c_i $
to ensure that the sum of the capacities is half of the total weight sum.
Degenerate instances were discarded as in Pisinger's experiments 
\citeyear{Pisinger99}.  That is, we only used instances where:
(a) each of the items fits into at least one of the containers, 
(b) the smallest container is large enough to hold at least the smallest item, and 
(c) the sum of the item weights is at least as great as the size of the
largest container.
In our experiments, we used items with weights in the range [10,1000].


\subsubsection{Comparison of Bin Completion With Previous Algorithms}

We now compare bin completion (BC) with Mulknap and MTM.
In all experiments described below involving Mulknap, 
we used Pisinger's Mulknap code, available at his web site\footnote{http://www.diku.dk/$\sim$pisinger/}, compiled using the {\texttt gcc}
compiler with the \texttt{-O3} optimization setting. 
Likewise, in all experiments described below involving MTM, we used Martello and
Toth's Fortran implementation of MTM from \cite{MartelloT90}, which was converted to C using \texttt{f2c} so that we could add some instrumentation.
Our bin completion code was implemented in Common Lisp.\footnote{All of the bin completion solvers described in this paper for the MKP, MCCP, bin covering, and bin packing problems are implemented in Common Lisp.}
We have shown experimentally that the choice of programming language added an overhead of approximately a factor of two to our runtimes (see \citeR{Fukunaga05}, Appendix A) compared to GNU C 2.95 with \texttt{-O3} optimization.



\begin{table}
  \begin{center}
\begin{footnotesize}
    \begin{tabular}{|r|r|r|r|r|r|r|r|r|r|r|r|r|r|r|r|r|}
	\hline
\multicolumn{2}{|c|}{} & \multicolumn{3}{|c|}{MTM} & \multicolumn{3}{c|}{Mulknap} & \multicolumn{3}{c|}{Bin Completion+NDP} \\
	\hline
m &  n & fail & time & nodes &   fail & time & nodes & fail & time & nodes \\
{\tiny \# bins} & {\tiny \# items} & & & & & & & & & \\
\hline
\multicolumn{11}{|c|}{{\bf Uncorrelated Instances}}\\
\hline
2&20 & 0 & 0.0003 & 201 &       0 & 0.0000    & 1   &    0& 0.0050 & 55 \\
3&20 & 0 & 0.0017 & 697 &       0 & 0.0000    & 23 &    0& 0.0063 & 123\\
4&20 & 0 & 0.0063 & 1530 &      0 & 0.0030 & 306 &   0& 0.0023 & 112\\
5&20 & 0 & 0.0767 & 34800 &     0 & 0.0387 & 2977&   0& 0.0033 & 212\\
6&20 & 0 & 0.2343 & 100387 &    0 & 0.4063 & 36057&  0& 0.0050 & 403\\
7&20 & 1 & 1.5721 & 851591 &    1 & 3.8207 & 369846& 0& 0.0043 & 399\\
8&20 & 1 & 6.3352 & 3539729 &   1 & 23.7996& 2493203& 0& 0.0023 & 207\\
9&20 & 1 & 18.3314 & 11294580 &   4 & 33.2138& 2868979& 0& 0.0017 & 136\\
10&20 & 3&22.5185 & 13536735 &  5 & 52.0812& 4848676& 0& 0.0010 & 84\\

\hline
\multicolumn{11}{|c|}{{\bf Weakly Correlated Instances}}\\
\hline
2&20 & 0 &0.0040 & 781 &        0 & 0.0000    & 26 &   0 & 0.0340 & 87\\
3&20 & 0 &0.0127 & 2693 &       0 & 0.0167 & 638 &  0 & 0.0117 & 240\\
4&20 & 0 &0.0327 & 6626 &       0 & 0.0390  & 1708&  0 & 0.0133 & 379\\
5&20 & 0 &0.2363 & 53324 &      0 & 0.2917 & 12701& 0 & 0.0107 & 521\\
6&20 & 0 &0.2763 & 65216 &      0 & 0.4220 & 20165& 0 & 0.0080 & 405\\
7&20 & 0 &0.6237 & 168974 &     0 & 1.3180 & 60890&  0 & 0.0050  & 245\\
8&20 & 0 &11.0927 & 2922499 &   2 & 6.5793 & 40339& 0 & 0.0027 & 144\\
9&20 & 0 &27.5807 & 11263928 &  3 & 22.5878& 1274756&0 & 0.0013 & 66\\
10&20 & 2&25.1029 & 10697165 &  5 & 72.3472&4478185& 0 & 0.0013 & 43\\
        \hline
\multicolumn{11}{|c|}{{\bf Strongly Correlated Instances}}\\
\hline
2&20 & 0 &0.0723 & 5747 &     0 & 0.0037 & 32&    0 & 0.0743 & 118\\
3&20 & 0 &0.1243 & 6748 &     0 & 0.0053 & 65&    0 & 0.0417 & 278\\
4&20 & 0 &0.1237 & 8125 &     0 & 0.0120 & 156&   0 & 0.0353 & 331\\
5&20 & 0 &0.1777 & 14823 &     0 & 0.0360 & 563&   0 & 0.0350 & 359\\
6&20 & 0 &0.1807 & 18012&     0 & 0.0550 & 734&  0 & 0.0217 & 194\\
7&20 & 0 &5.1393 & 882080 &    1 & 0.0928 & 1339&  0 & 0.0137 & 109\\
8&20 & 2 &11.2361 & 2616615&   3 & 12.2581& 225236&0 & 0.0073 & 42\\
9&20 & 2 &20.0136 & 3150848&   4 & 33.5415& 693398& 0 & 0.0060 & 25\\
10&20& 4 &38.8554 & 8141902&   9 & 34.3291 & 634358&0 & 0.0057 & 21\\

\hline
\multicolumn{11}{|c|}{{\bf Subset-Sum Instances}}\\
\hline
2&20 & 0 &0.0820 &2061 &   0 & 0.0033 & 9 &      0 & 0.0447 & 64\\
3&20 & 0 &0.2057 & 7033 &    0 & 0.0337 & 77&      0 & 0.0310 & 248\\
4&20 & 0 &0.1760 & 8371 &    0 & 0.0503 & 137&     0 & 0.0260 & 294\\
5&20 & 0 &0.2507 & 17514 &    0 & 0.1780 & 830&     0 & 0.0257 & 315\\
6&20 & 0 &0.2393 & 17546&    0 & 0.1953 & 738&     0 & 0.0173 & 176\\
7&20 & 0 &5.7657 & 852615&    1 & 0.3221 & 1752&    0 & 0.0113 & 102\\
8&20 & 2 &14.9261& 2546834&   3 & 19.5930& 270841&  0 & 0.0073 & 39\\
9&20 & 2 &25.9393 & 3124421&  4 & 41.8962& 677467&  0 & 0.0063 & 25\\
10&20 &4 &51.5027 & 7907407&  9 & 39.7500 & 627688&  0 & 0.0057 & 20\\
        \hline
    \end{tabular}
\end{footnotesize}
  \end{center}

  \caption{MTM, Mulknap, and Bin Completion with Nogood Dominance Pruning on small MKP instances with varying $n/m$ ratio.The {\em fail} column indicates the 
number of instances (out of 30) that were not solved within the time limit (300 seconds/instance). The {\em time} (seconds on 2.7GHz Pentium 4) and {\em nodes} columns
show average time spent and nodes generated on the
successful runs, excluding the failed runs.}

  \label{tab:mtm-mulknap-bc-20-1}
\end{table}

For each of the four problem classes, we generated test sets of 30 instances each, where $n=20$, and $m$ was varied between 2 and 10.
On each instance, we ran Mulknap, bin completion, and bin completion
with nogood dominance pruning.  
The purpose of this experiment was to observe the behavior of the algorithms as the ratio $n/m$ was varied.
The results are shown in Table \ref{tab:mtm-mulknap-bc-20-1}.
Each algorithm was given a time limit of
300 seconds to solve each instance.  The {\em fail} column indicates
the number of instances (out of 30) which could not be solved by the
algorithm within the time limit.  The {\em time} and {\em nodes}
column show the average time spent and nodes generated on the
successful runs, excluding the failed runs.
All experiments described in this section were run on a 2.7GHz Pentium 4.

Our data confirms the observations of \citeA{Pisinger99} and \citeA{MartelloT81} that
the class of uniform,
random instances that require the most search for previous branch-and-bound
solvers appear to be generated when $n/m$ is relatively low.
In other words, the $n/m$ ratio for the MKP appears to be a critical parameter that determines search difficulty, similar to the clause-to-variable ratio for satisfiability problems \cite{MitchellSL92}.
Table \ref{tab:mtm-mulknap-bc-20-1} shows
that Mulknap and MTM have great difficulty with relatively small
problems with small $n/m$ ratios, while bin completion could solve
these problems very quickly.

Next, we investigated larger problem instances, where $n/m$ varied between 2 and 4.
In this experiment, we also included bin completion with nogood pruning (BC+NP), in addition to MTM, Mulknap, and BC+NDP. 

As shown in Table \ref{tab:mkp-results-1}, for problems where $n/m$ =  2 or 3, 
both variants of bin completion (BC+NP and BC+NDP) dramatically outperformed both MTM and Mulknap, 
with the difference in performance becoming more pronounced
as problem size was increased. 
Because the runs were truncated at 300 seconds per instance per
algorithm, it is not possible to compare the full runtimes for all of
the sets. Note, for example, that on subset-sum instances
where $n=40$ and $m=20$, the mean runtime for BC+NDP is 0.06 seconds,
while neither MTM nor Mulknap solved any of the instances within 300
seconds each, so they are both {\em at least} three orders of magnitude
slower than BC+NDP on these instances.
We tried running the solvers longer in order to get actual runtimes
(as opposed to a truncated, lower bound on runtimes), but found that
even allocating an hour per problem was insufficient to allow MTM and
Mulknap to solve any of these problems.
These results suggest that bin completion is
asymptotically more efficient than MTM and Mulknap for this class of problem instances.
For the problems with $n/m$ = 4, the results are similar, with the
notable exception that for the strongly-correlated instances, Mulknap
outperformed bin completion when $n=40$ and $m=10$.

For problems where $n/m$ = 2 and 3, we observed that BC+NDP
consistently outperforms BC+NP by a significant margin with respect to
the number of nodes searched, and significant improvements in success
rate and execution times are observed for the larger problems sets.
However, when $n/m$ = 4, BC+NDP still searches fewer nodes than BC+NP,
but the difference is much less significant, and in fact, the reduced
search is not enough to offset the increased overhead per node for NDP,
so that the runtimes of BC+NP and BC+NDP end up being comparable (Table \ref{tab:mkp-results-1}).

We also ran some experiments with instances with larger $n/m$ ratios.
When $n/m \geq 5$, Mulknap clearly dominates all other current
algorithms, solving instances with little or no search in under a
second, as first demonstrated by \citeA{Pisinger99}.  At high
$n/m$ ratios, the bound-and-bound approach of Mulknap has a very high
probability of finding an upper bound that matches the lower bound at
each node, essentially eliminating the need for search. In fact,
\citeA{Pisinger99} showed that for instances with $n/m \geq 10$, there were
almost no instances that required searching more than one node.


%

\begin{landscape}

\begin{table}
  \begin{center}
\begin{scriptsize}
    \begin{tabular}{|r|r|r|r|r|r|r|r|r|r|r|r|r|r|}
	\hline
\multicolumn{2}{|c|}{} &  \multicolumn{3}{c|}{MTM} &\multicolumn{3}{|c|}{Mulknap    } & \multicolumn{3}{|c|}{Bin Completion+NP} & \multicolumn{3}{c|}{Bin Completion + NDP}  \\
	\hline
m &  n 	& fail & time & nodes & fail & time & nodes & fail & time & nodes &  fail & time & nodes  \\
\# bins & \# items & & & & & & & & & & & & \\
\hline
\multicolumn{14}{|c|}{{\bf Uncorrelated Instances}}\\
\hline
15&30 & 30 & -     & -          & 30 & - & - &               0 & 0.04 & 1215 &  0 & 0.02 & 887 \\
20&40 & 30 & - & -              & 30 & - & - &               0 & 1.48 & 66681 & 0 & 0.72 & 26459 \\
25&50 & 30 & - & -              & 30 & - & - &               4 & 72.66& 2767939 & 3 & 32.43 & 1060723\\
10&30 & 12& 47.77 & 9240426     & 11  & 29.96 & 1649062 &  0 & 0.71 & 48407   & 0 & 0.61 & 33141\\
15&45 & 30 & - & -              & 30  & - & - &              15 & 66.81 & 3369997 & 15 & 60.54 & 2516681\\
 5&20 & 0 & 0.08 & 32602 &         0  & 0.10 & 8487 &        0  & 0.01 & 253    & 0 & 0.01 & 237 \\
10&40 & 7 & 99.13 & 15083247 &  4 & 51.66 & 1744911 &     0  & 11.58 & 360567 & 0 & 13.47 & 297225 \\
\hline
\multicolumn{14}{|c|}{{\bf Weakly Correlated Instances}}\\
\hline
15&30 & 30 & -     & -          & 30 & - & - &               0  & 0.07 & 2123 &   0  & 0.05 & 1050 \\
20&40 & 30 & -     & -          & 30 & - & - &               0  & 0.79 & 22553 &  0  & 0.35 & 7473 \\
25&50 & 30 & -     & -          & 30 & - & - &               0  & 34.39 & 691645 & 0 & 15.80 & 231934\\
10&30 & 16 & 68.80 & 8252155    & 15 & 69.65 & 1647576 &   0  & 0.94 & 34001  &  0  & 0.79  & 25764\\
15&45 & 30 & - & -              & 30 & - & - &               19 & 52.42 & 1126242 & 18 &67.17 & 1207099\\
 5&20 &  0 & 0.10 & 21150       &  0 & 0.14 & 5758 &       0  & 0.01 &  407    & 0  & 0.01 & 396 \\
10&40 & 29 & 205.99 & 12672965 & 27 & 131.41 & 2390152 &  0  & 21.10 & 565036 & 0  & 20.63 & 490983\\
	\hline


\hline
\multicolumn{14}{|c|}{{\bf Strongly Correlated Instances}}\\
\hline
15&30 & 30 & -     & -          & 30 & - & - &               0 & 0.04 & 415      & 0 & 0.03 & 194\\
20&40 & 30 & -     & -          & 30 & - & - &               0 & 3.46 & 5407     & 0 & 2.29 & 2694\\
25&50 & 30 & -     & -          & 30 & - & - &               9 & 52.26 & 58500   & 4 & 58.72 & 61900\\
10&30 &  1 & 23.48 & 2916858 &     0 & 7.02 & 88737 &        0 & 0.53 & 5662     & 0 & 0.50 & 5381\\
15&45 & 30 & - & -              & 30 & - & - &              24 & 105.90 & 964566 & 23 & 118.72 & 1150024\\
 5&20 &  0 &  0.06 & 12490 &       0 & 0.02  & 486 &           0 & 0.01 & 264      & 0 & 0.01 & 258\\
10&40 & 13 & 115.38 & 5528847 &    0 & 44.55 & 361987 &      2 & 51.89 & 263462 &  2 & 50.51 & 254094\\

\hline
\multicolumn{14}{|c|}{{\bf Subset-Sum Instances}}\\
\hline
15&30 & 30 & -     & -          & 30 & - & - &               0 & 0.01 & 171      & 0 & 0.01 & 95\\
20&40 & 30 & -     & -          & 30 & - & - &               0 & 0.13 & 1664     & 0 & 0.06 & 806\\
25&50 & 30 & -     & -          & 30 & - & - &               0 & 4.37 & 55334    & 0 & 1.59 & 19491\\
10&30 & 0 & 25.58 & 1993648 &      0 & 4.74 & 43734 &        0 & 0.12 & 1816     & 0 & 0.12 & 1752\\
15&45 & 18 & 0.0 & 27.58 &        23 & 100.19 & 480014 &     2 & 45.07 & 597581  & 2 & 43.44 & 562129\\
 5&20 & 0 & 0.05 & 5676 &          0 & 0.01   & 207 &     0 & 0.01 & 140      & 0 & 0.01  & 137 \\
10&40 & 5 & 35.20 & 946934 &       0 & 10.48  & 58317  &     0 & 1.98 & 26674    & 0 & 2.00 & 26097 \\
	\hline
    \end{tabular}
\end{scriptsize}
  \end{center}
  \caption{Multiple knapsack problem results: Comparison of MTM, Mulknap, Bin Completion with Nogood Pruning (NP), and Bin Completion with Nogood Dominance Pruning (NDP) on hard MKP instances.
The {\em fail} column indicates the
number of instances (out of 30) that were not solved within the time limit (300 seconds/instance). The {\em time} (seconds on 2.7GHz Pentium 4) and {\em nodes}
show average time spent and nodes generated on the
successful runs, excluding the failed runs.}
  \label{tab:mkp-results-1}
\end{table}

\end{landscape}

On the other hand, for $n/m \geq 5$, bin completion tends to generate a very large (more than 10000) number
of undominated bin assignments at each decision node, and
runs out of memory while allocating the set of undominated bin
assignments that are the candidates to be assigned to the current bin. 
While hybrid incremental branching (Section \ref{sec:hybrid-incremental}) eliminates
this problem and allows the algorithm to run to completion, it is
still not competitive with Mulknap (runs do not complete within a 300-second time limit).

We conclude that uniform instances of the multiple knapsack problem exhibit a ``bimodal''
characteristic, for which bin completion  and bound-and-bound are complementary approaches.
For $n/m \leq 4$, uniform MKP instances require a significant amount
of search to solve, and our bin completion approach is clearly the current
state of the art, as demonstrated by 
Table \ref{tab:mkp-results-1}.
On the other hand, for $n/m \geq 5$, the bound-and-bound approach as
exemplified by Pisinger's Mulknap algorithm is the state of the art,
and the runtimes are dominated by the computation of a single lower
bound and a single upper bound at the root node. 
There is a rather sharp ``phase transition'' around $n/m = 4$ where
the dominant approach changes between bin completion and bound-and-bound 
(see \citeR{Fukunaga05} for details).


Because of the highly complementary nature of Mulknap and
bin completion, the MKP is a natural candidate for the application of
an {\em algorithm portfolio} approach \cite{HubermanLH97,GomesS01} where both
Mulknap and bin completion are run in parallel on the same problem instance. Even under a trivial
portfolio resource allocation scheme where both processes received
equal time, the resulting total CPU usage of the
portfolio on each instance would be no worse than twice that of the faster algorithm.
Another way to combine bin completion with bound-and-bound is to add bound-and-bound at each node of the bin completion search, and is an avenue for future work.

 
 
 

\section{Min-Cost Covering Problem (MCCP)}
\label{sec:min-cost-covering}

Given a set of containers with quotas and a set of items (characterized by a
weight and cost), the objective of the MCCP is
to assign items to containers such that all the container quotas are
satisfied, and the sum of the costs of the items assigned to the
containers is minimized (see \ref{sec:mccp-definition} for a formal definition).

\subsection{Christofides, Mingozzi, and Toth (CMT) Algorithm}

The previous state-of-the-art algorithm for the min-cost covering
problem is an early version of bin completion by Christofides,
Mingozzi, and Toth \citeyear{ChristofidesMT79}.  
Their algorithm is a bin completion algorithm which uses the CMT dominance criterion (see Section \ref{sec:dominance-criteria}).

An effective lower bound can be computed by solving $m$ independent
minimization problems (similar to the standard 0-1 Knapsack, except that the objective is to satisfy a container's quota while minimizing the cost of the items assigned to it), one for each of the bins, and then summing the
results. This $L_2$ lower bound is a relaxation of the constraint that each item can
only be used once. 




\citeA{ChristofidesMT79} proposed several more complex lower bounds, but
they found empirically that among all of the proposed lower bounds,
the $L_2$ bound resulted in the best performance by far across a wide
range of problem instances.


\subsection{Bin Completion for the MCCP}

Our new bin completion algorithm for the MCCP is similar to the
Christofides et al. algorithm. The major difference is that we use a more powerful dominance criterion (Proposition \ref{def:mccp-dominance-criterion}).
Each node in the depth-first, search tree represents a minimal, feasible bin assignment for a particular bin. 
Our bin completion algorithm assigns bins in order of non-decreasing size, i.e., a smallest-bin-first variable ordering heuristic. 
We evaluated eight different strategies for ordering the candidate undominated bin assignments, and found that a {\em
min-weight} strategy which sorts assignments in non-decreasing
order of weight performed best. Fukunaga \citeyear{Fukunaga05} provides a detailed comparison of variable and value orderings for the MCCP.
We use the same $L_2$ bound as the CMT algorithm.

\subsection{Experimental Results}


 We compared the performance of bin completion variants and previous algorithms.
We implemented the following algorithms.

\begin{itemize}
\item \texttt{CMT} - The Christofides, Mingozzi and Toth algorithm described above.
Our implementation used smallest-bin-first variable ordering, min-weight value ordering, and the $L_2$ lower bound.
\item \texttt{CMT+NP} - The CMT algorithm extended with nogood pruning.
\item \texttt{BC} - Bin completion using min-weight value ordering and smallest-bin-first variable ordering.
\item \texttt{BC+NP} - Bin completion with nogood pruning
\item \texttt{BC+NDP} - Bin completion with nogood dominance pruning, 
\end{itemize}

We used smallest-bin-first variable ordering for the CMT and BC
variants after observing that this led to good performance compared
with random or largest-bin-first variable orderings.

In the experiment shown in Table \ref{tab:mccp-results1},
we compared CMT, CMT+NP, BC, BC+NP, and BC+NDP.
The purpose of this experiment was to evaluate the relative impact of each
component of bin completion, by comparing various combinations of: (1) the type of dominance
criterion used, (2) whether nogood pruning was used, and (3) whether
nogood dominance pruning was used.

We used the same four classes of test problems (uncorrelated, weakly correlated, strongly correlated, and subset-sum) as for the multiple knapsack problem experiments (Section \ref{sec:mkp-results}), with item weights and costs in the range [10,1000].
For each of the four problem classes, 30
instances were generated for various values of $m$ and $n$.
We ran each algorithm on each instance.
%
All
experiments were run on a 2.7GHz Pentium 4.  Each algorithm was given
a time limit of 300 seconds to solve each instance.  The {\em fail}
column indicates the number of instances (out of 30) which could not
be solved by the algorithm within the time limit.  The {\em time} and
{\em nodes} column show the total time spent and nodes generated, excluding the failed runs.

\begin{landscape}

\begin{table}
  \begin{center}
\begin{scriptsize}
    \begin{tabular}{|r|r|r|r|r|r|r|r|r|r|r|r|r|r|r|r|r|r|r|}
	\hline
\multicolumn{2}{|c|}{} &  \multicolumn{3}{c|}{CMT} & \multicolumn{3}{c|}{CMT+NP} & \multicolumn{3}{c|}{BC} & \multicolumn{3}{c|}{BC+NP} & \multicolumn{3}{c|}{BC+NDP}  \\
	\hline
 m&n 	& {\bf fail} & time & nodes & {\bf fail} & time & nodes &   {\bf fail} & time & nodes & {\bf fail} & time & nodes & {\bf fail} & time & nodes \\
\# bins & \# items & & & & & & & & & & & & & & &\\
\hline
\multicolumn{17}{|c|}{{\bf Uncorrelated Instances}}\\
\hline
5&15  & 0  & 0.30 & 34094 &  0 & 0.21 & 12641 & 0 & 0.02 & 678        & 0 & 0.01 & 385      &  0 & 0.01 & 238\\
10&30 & 30 & - & -& 30 & - & -                & 20& 184.79 & 12816074 & 2 & 49.66 & 1952236 & 0 & 23.82 & 635035\\
5&10  &  0 & 0.01 & 256 &   0 & 0.01 & 48     & 0 & 0.01 & 127        & 0 & 0.01 & 31       & 0 & 0.01 &  28\\
10&20 & 1 & 47.43 & 5136873 & 0 & 0.17 & 10645 &0 & 7.24 & 801863     & 0 & 0.02 & 2149     & 0 & 0.01 & 1149\\
15&30 & 30 & - & -          & 11 & 80.04 & 3121178 & 30 & - & -       & 0 & 11.58 & 705748  & 0 & 2.81 & 136000\\
20&40 & 30 & - & -          & 30 & - & - & 30 & - & -                 & 26 & 184.59 & 8356961& 10& 97.67 & 3145263\\
5&20  & 0 & 13.66 & 1041966 & 0  & 0.10 & 4665 & 0 & 0.29 & 15979     & 0 & 0.25  & 9339    & 0 & 0.15 & 5193\\
\hline
\multicolumn{17}{|c|}{{\bf Weakly Correlated Instances}}\\
\hline
5&15  & 0 & 0.21 & 10239   & 0 & 0.13 & 4665     & 0 & 0.01 & 347       & 0 & 0.01 & 197   & 0 & 0.01 & 150\\
10&30 & 30 & - & -         & 30 & - & -      & 26 & 208 & 4242634   & 16 & 168.21 & 2843080 & 5 & 97.16 & 1436142\\
5&10  & 0 & 0.01 & 46      & 0 & 0.01 & 21   & 0 & 0.01 & 25        & 0 & 0.01 & 12    & 0 & 0.01 & 12\\
10&20 & 0 & 3.22 & 189874  & 0 & 0.07 & 3979 & 0 & 0.13 & 11216     & 0 & 0.01 & 428   & 0 & 0.01 & 236\\
15&30 & 28 & 227.63 & 16471570 & 2 & 46.49 & 1803180 & 20 & 106.70 & 5367215 & 0 & 3.19 & 182830 & 0 & 0.56 & 30669\\
20&40 & 30 & - & -         & 18 & 97.58 & 2874351 & 30 & - & -      & 12 & 112.08 & 3871911 & 2 & 44.84 & 1704351\\
5&20  & 1 & 54.11 & 975879 & 0 & 17.64 & 203251 & 0 & 0.65 & 15566  & 0 & 0.54 & 10032 & 0 & 0.45 & 7698\\

	\hline



\multicolumn{17}{|c|}{{\bf Strongly Correlated Instances}}\\
\hline
5&15  & 0 & 0.08 & 4226  & 0 & 0.05 & 2518 & 0 & 0.01 & 68 & 0 & 0.01 & 53 & 0 & 0.01 & 40\\
10&30 & 30 & - & -       & 30 & -   & -    &  0   & 54.65 & 1340276 & 0 & 17.60 & 333450 & 0 & 6.20 & 114786\\
5&10  & 0 & 0.01 & 30 & 0 & 0.01 & 16 & 0 & 0.01 & 10 & 0 & 0.01 & 7 & 0 & 0.01 & 7 \\
10&20 & 0 & 0.24 & 19496 & 0 & 0.02 & 1064 & 0 & 0.01 & 1064 & 0 & 0.01 & 139 & 0 & 0.01 & 79 \\
15&30 & 6 & 93.33 & 5178573 & 0 & 5.18 & 241654 & 0 & 2.45 & 204069 0 & 0 & 0.12 & 7532 & 0 & 0.02 & 1354 \\
20&40 & 30 & - & - & 18 & 97.19 & 2844301 & 17 & 52.50 & 2933474 & 1 & 27.17 & 1039810 & 0 & 2.13 & 86685\\
5&20  &  0 & 18.91 & 446419 & 0 & 20.78 & 253291 & 0 & 0.09 & 2474 & 0 & 0.07 & 1893 & 0 & 0.06 & 1546\\

\hline
\multicolumn{17}{|c|}{{\bf Subset-Sum Instances}}\\
\hline
5&15  & 0 & 0.08 & 4606 & 0 & 0.07 & 2899 & 0 & 0.01 & 65            & 0 & 0.01 & 49  & 0 & 0.01 & 42\\
10&30 & 30 & - & - & 30 & - & -           & 0 & 32.71 & 810471       & 0 & 11.32 & 213975 & 0 & 3.97 & 70774\\
5&10  & 0 & 0.01 & 31 & 0 & 0.01 & 17     & 0 & 0.01 & 10            & 0 & 0.01  & 8  & 0 & 0.01 & 8\\
10&20 & 0 & 0.17 & 13286 & 0 & 0.03 & 1573 & 0 & 0.01& 986           & 0 & 0.01 & 146 & 0 & 0.01 & 73\\
15&30 & 10 & 74.57 & 3812510 & 0 & 17.48 & 789928 & 0 & 2.34 & 177683  & 0 & 0.10 & 6168 & 0 & 0.02 & 1272\\
20&40 & 30 & - & - & 18 & 97.90 & 2874340 & 17 & 55.18 & 2933474                & 2 & 26.37 & 1160452 & 0 & 2.47 & 125697\\
5&20  & 0 & 18.11 & 384584 & 0 & 17.66 & 203251 & 0 & 0.08 & 2270    & 0 & 0.07 & 1605  & 0 & 0.06 & 1242\\
\hline
    \end{tabular}
\end{scriptsize}
  \end{center}
  \caption{{\footnotesize Min-cost covering problem results: Comparison of (a) Christofides, Mingozzi, and Toth (CMT) algorithm,  (b) CMT algorihm with our Nogood Pruning (NP), (c) Bin Completion, (d) Bin Completion with Nogood Pruning (NP), and (e) Bin Completion with Nogood Dominance Pruning (NDP).
The {\em fail} column indicates the
number of instances (out of 30)  that were not solved within the time limit. The {\em time} (seconds on 2.7GHz Pentium 4) and {\em nodes} columns
show average time spent and nodes generated on the
successful runs, excluding the failed runs.}}

  \label{tab:mccp-results1}
\end{table}

\end{landscape}

As shown in 
Table \ref{tab:mccp-results1},
each component of bin completion has a
significant impact.  Although our dominance criterion requires much
more computation per node  than the simpler CMT criterion,
the search efficiency is dramatically improved. Thus, BC performed much
better than CMT, and BC+NP performed much better than CMT+NP.  

Furthermore, the nogood pruning (NP) strategy significantly improves performance for
both the algorithm based on the CMT dominance criterion, as well as
our dominance criterion, as evidenced by the improvement of CMT+NP
over CMT and the improvement of BC+NP over BC. In fact, nogood pruning
is sufficiently powerful that it allows CMT+NP to sometimes outperform
pure BC (without nogood pruning). Thus, this data illustrates the power of nogood pruning, confirming similar results for bin packing reported in \cite{Korf03}, as well as in some preliminary experiments for the MKP and MCCP.

Finally, BC+NDP results in the
best performance, significantly outperforming BC+NP for the larger instances with respect to the number of problems solved within the time limit, as well as the runtimes and nodes on the problems that were solved.


We also implemented two baseline algorithms: a
straightforward integer programming model using the freely available GNU {\tt glpk} integer programming solver, as well as an item-oriented
branch-and-bound algorithm which uses the $L_2$ lower bound, but these
baselines performed very poorly compared to the CMT and bin completion
algorithms \cite{Fukunaga05}.

\section{The Bin Covering Problem}
\label{sec:bin-covering}

Given a set of identical containers with quota $q$ and a set of $n$ items, each with weight $w_i$, the bin covering problem, sometimes
called {\em dual bin packing}, is to assign all items to containers such that the number of
containers whose quotas are satisfied (i.e., the sum of item weights
assigned to the container equal or exceed the quota) is maximized (see \ref{sec:bc-definition} for formal definition).
Although there has been considerable interest in the bin covering
problem in the algorithm and operations research communities, most of
the previous work on bin covering has been theoretical, focusing on
approximation algorithms or heuristic
algorithms \cite<e.g.,>{AssmannJKL84,FosterV89,CsirikFGK91,CsirikJK01}, and analysis of the
properties of some classes of instances \cite<e.g.,>{Rhee89}.

\subsection{The Labb\'{e}, Laporte, Martello Algorithm}

The state-of-the-art algorithm for opimally solving bin covering is an
item-oriented branch-and-bound algorithm by Labb\'{e}, Laporte, and
Martello \citeyear{LabbeLM95}. We refer to this as the {\em LLM}
algorithm.
The items are sorted in non-increasing order of size.  Each node
represents a decision as to which bin to put an item into.  At each
node, upper bounds based on combinatorial arguments are computed, and
the remaining subproblem is reduced using two reduction criteria. At
the root node, a set of heuristics is applied in order to compute an
initial solution and lower bound.
The LLM upper and lower bounds are described in \cite{LabbeLM95}.

\subsection{Bin Completion for Bin Covering}

Our bin completion algorithm for bin covering works as follows.
First, we use the LLM upper bounding heuristics at the root node
to find an initial solution and lower bound.
Then, we apply a bin completion branch-and-bound
algorithm, using our new dominance criterion (Proposition \ref{def:bc-dominance-criterion}).  
Each node in the depth-first, search tree represents a minimal, feasible bin assignment for a particular bin that include the largest remaining item. 
At each node, we apply the same upper bounding procedures as the LLM algorithm
to compute an upper bound.
In addition, we apply the LLM reduction criteria at each node.
We evaluated eight different strategies for ordering undominated bin assignments, and found that the min-cardinality-min-sum strategy (sort bin assignments in order of non-decreasing cardinality, breaking ties by non-decreasing sum) performed best overall \cite{Fukunaga05}.


\subsection{Empirical Results}

In order to evaluate our bin covering algorithm, we considered the class of uniform,
random problem instances 
previously studied by Labb\'{e}, Laporte, and Martello.
This is a simple model in which $n$ items are chosen uniformly in the
 range $[min,max]$, where $max$ is less than the bin
 quota $q$.
In their experimental evaluations, Labb\'{e}, Laporte, and Martello used items with weights in the range 1 and 100, and bin quotas ranging between 100 and 500.

However, many instances in this class can be solved without any
search. If the LLM lower bound heuristics find a solution with the same number of bins as the LLM upper bound, then we have found the optimal solution and can terminate without any search. 
We say that an instance is {\em trivial} if it is solved without any
search, and {\em nontrivial} otherwise.

To illustrate the prepondrance of trivial instances,
we generated 10000 uniform, random instances with bin quota 100000 and 120 items with weights in the range $[1,99999]$. Of these, 9084 were solved at the root node. This shows that most uniform random bin covering instances are, in fact, trivial given powerful upper and lower bounds.
We have previously observed a similar phenomenon for bin packing -- most
uniform, random bin packing instances are solved at the root node by the
combination of the best-first-decreasing heuristic and the
Martello-Toth L2 lower bound \cite{Korf02}.


It is well-known that the number of significant digits of precision in
the weights of the items significantly affects the difficulty of
one-dimensional packing problems such as the 0-1 Knapsack problem
\cite{KellererPP04}. In general, problem difficulty increases with
precision. This property extends to multicontainer, one-dimensional packing problems as well \cite<e.g.,>{Pisinger99}.
We confirmed that problem difficulty was highly correlated with the number of significant digits of item weights \cite{Fukunaga05}.



Therefore, in the experiments described below, we only used nontrivial instances, in
order to highlight the impact of differences in search strategy.  That
is, when generating test instances, we filtered out all trivial instances by
testing whether the LLM upper bound is matched by its lower bound
heuristics.
Furthermore, we use high precision problems (quotas of 10000 or more) in order to focus on the most difficult instances.




\subsubsection{Hybrid Incremental Branching}
\label{sec:bin-covering-hybrid-incremental}
In Section \ref{sec:hybrid-incremental}, we proposed hybrid
incremental branching, a strategy for avoiding the runtime and memory
overheads imposed by completely enumerating all undominated children
of a node.
In our earlier experiments with the multiple knapsack and min-cost
covering problems, (Sections
\ref{sec:multiple-knapsack}--\ref{sec:min-cost-covering}), the problem
instances used in our experiments had fewer than 50 items, and
we have not observed more than
a few hundred assignments generated at each node in our MKP and MCCP experiments.
Thus, the number of candidate undominated bin assignments generated per node
did not become a bottleneck,
and hybrid incremental branching was unnecessary.
%
However, as (1) the average number of items that fit in a container
increases, and (2) the number of items increases, 
the number of
candidate undominated bin assignments per
node increases. Therefore, the benchmark comparisons below for bin covering, hybrid incremental branching becomes much more relevant.

To illustrate this, we performed the following experiment.  We
generated 20 nontrivial instances with bin quota $q$=20000 and 100
items in the range [1,9999].  We applied bin completion + NDP to these
problem instances, using hybrid incremental branching with various
values for the parameter $h$, which limits the number of children that are generated at once at every node. 
For $h$= 2000, 200, 20, and 2, the instances were solved on average in 4.998 seconds, 0.150 seconds, 0.0139 seconds, and 0.0079 seconds, respectively.
The average number of nodes expanded was 20.2 for all $h \in \{2,20,200,2000\}$.
In other words, these instances were easily solved by relying on the leftmost children at each node in the bin completion search tree -- generating additional children was an entirely unnecessary but very expensive overhead.
Thus, as $h$ increases, the same number of nodes are being
explored, but each node requires much more computation to enumerate and sort up to $h$ undominated children, resulting in
more than two orders of magnitude difference in runtime between $h=2$
and $h=2000$. 
To see how large $h$ would have to be in order to be able to enumerate
{\em all} of the undominated children of a node, we have experimented
with $h$ up to 10000, but found it insufficient (the statically
allocated array of size $h$ overflowed during the runs).

We experimented with several values of $h$ on several classes of
problems, but found that the optimal value of $h$ varied significantly
depending on the problem instance.  In our experiments below, we use
hybrid incremental branching with $h$=100, in order to set a
balance between minimizing the number of nodes expanded (exploiting value ordering among the children of each node) and minimizing
the computational overhead per node (by minimizing the number of children generated).

\subsubsection{Comparing LLM and Bin Completion}

Next, we compared LLM, bin completion+NP, and
bin completion+NDP 
using larger instances. 
We also implemented a straightforward integer linear programming model using GNU {\tt glpk} for bin covering, but found that it performed much worse than the LLM and bin completion algorithms \cite{Fukunaga05}.

%
For each $n \in \{60,80,100\}$,
we generated 2000 non-trivial, uniform random instances where the items were chosen
uniformly in the range [1,99999], and the bin quota was 100000. 
We also generated 2000 non-trivial instances where $n=100$,  $q= 200000$, and items were in the range [1,99999].
We ran our
implementations of the three algorithms on 
each instance, with a time limit of 180 seconds per instance.
%
As shown in Table \ref{tab:dbp-results},
the bin completion algorithms significantly outperformed LLM. 
On the harder problems, bin completion + NDP significantly outperforms bin completion + NP.
For the problems with $n=100$, $q=200000$, our results indicate that
while these instances are extremely difficult for the LLM algorithm, which failed to solve any of the 100 instances within the time
limit, these problems are actually relatively easy for
bin completion. 
Since little or no search is being performed by bin completion, bin completion+NDP does not improve upon bin completion+NP.

\begin{table}
  \begin{center}
\begin{small}
    \begin{tabular}{|r|r|r|r|r|r|r|r|r|r|r|r|r|r|r|}
	\hline
	&  &   \multicolumn{3}{c|} {Labb\'{e} et al.} & \multicolumn{3}{c|}{Bin Completion+NP} & \multicolumn{3}{c|}{Bin Completion+NDP}  \\
	\hline
n (\# of items) & q (quota)	& fail & time & nodes   	& fail & time & nodes &  fail & time & nodes  \\
\hline
     \multicolumn{11}{|c|}{{\bf 2000 nontrivial instances per set}}\\
\hline
      60 & 100000 & 10 & 0.39 & 25196 & 6 & 0.08 & 33287 & 2 & 0.11 & 13735\\
      80 & 100000 & 36 & 1.02 & 62093 & 15 & 0.18 & 39740 & 12 & 0.09 & 10517\\
     100 & 100000 & 45 & 1.98 & 113110 & 15 & 0.42  &152113 & 8 & 0.31 & 31107\\

\hline
     \multicolumn{11}{|c|}{{\bf 100 nontrivial instances per set}}\\
\hline

     100 & 200000 & 100 & - & - & 3 & 0.11 & 25 & 3 & 0.11 & 25\\
\hline
    \end{tabular}
\end{small}
  \end{center}
  \caption{Bin Covering results: Comparison of Labb\'{e}, Laporte, and Martello algorithm, Bin Completion with Nogood Pruning, and Bin Completion with Nogood Dominance Pruning. 
The {\em fail} column indicates the
number of instances (out of 30)  that were not solved within the time limit. The {\em time} (seconds on 2.7GHz Pentium 4) and {\em nodes} columns
show average time spent and nodes generated on the
successful runs, excluding the failed runs.}
  \label{tab:dbp-results}
\end{table}


\section{Bin Packing}
\label{sec:bin-packing}

The bin completion approach was originally developed for the
bin packing problem, and was shown to significantly outperform the Martello-Toth Procedure on randomly generated instances \cite{Korf02,Korf03}.
We implemented an extended bin completion algorithm for the bin
packing problem, and summarize our results below. For further details,
see \cite{Fukunaga05}.
Our bin completion based solver incorporated nogood dominance pruning and
a {\em min-cardinality-max-weight} value ordering strategy, in which
(1) completions are sorted in order of non-decreasing cardinality, and (2)
ties are broken according to non-increasing weight.

We evaluated our 
bin completion solver using the set of
standard OR-LIB instances.\footnote{available at
http://www.brunel.ac.uk/depts/ma/research/jeb/info.html}
This test
set consists of 80 ``triplet'' instances (where the elements are
generated three at a time so that the sum of the elements add exactly
to the bin capacity) and 80 ``uniform'' instances (where items sizes
are chosen from a uniform random distribution).  The bin completion
solver 
(executed on a 1.3GHz Athlon) was given a time limit of 15 minutes on
each instance from the benchmark set. It solved all 80 of the triplet
instances from OR-LIB within 1500 seconds combined. It solved all 40
of the uniform instances with 120 and 250 items in under 4 seconds
combined. It solved 18 (out of 20) of the 500-item uniform instances
in 11.13 seconds combined, but failed to solve 2 instances. Bin
completion solved 19 (out of 20) of the 1000-item uniform instances in
44 seconds combined, but failed to solve one of the instances.

However, 
our bin completion solver was not competitive
with the state of the art, 
which is a recently developed
branch-and-price integer linear programming solver by Belov and
Scheithauer \citeyear{BelovS06}.
Belov and Scheithauer 
provide a new benchmark set of
28 very hard bin packing instances; their solver solved most of these within
seconds, although some took hours.
Our current bin completion code could not solve any of the 28
instances, given 15 minutes per instance.
They also report that the largest triplet instances from OR-LIB
(Triplet-501) were solved in an average of 33 seconds per instance
(660 seconds total) on a 1GHz AMD Athlon XP. 
Furthermore, Belov was kind enough to run their solver on a set of one hundred,
80-item, uniform, random instances that we generated with items in the range
[1,1000000] and a bin capacity of 1000000.  Their solver
solved all of these instances in less than 1 second each at the root
node (i.e., without any search), using rounding heuristics based on
the linear programming LP solution, whereas our best bin completion
solver required 534 seconds and searched 75,791,226 nodes.

\subsection{Branch-and-Price vs. Bin Completion}

Recent branch-and-price approaches for bin packing such as the solver
by Belov and Scheithauer  use a bin-oriented branching strategy, where
decisions correspond to the instantiation of one or more maximal
bin assignments (see \citeR{ValeriodeCarvalho02}, for a
survey of branch-and-price approaches). 
At each node, a column generation procedure is used to compute
the LP lower bound.  They derive much of their power from a very
accurate LP lower bound based on a cutting-stock formulation of bin
packing \cite{GilmoreG61}, which has been observed to almost always
give the optimal value as the lower bound, and has never been observed
to give a value that is more than one greater than the optimal value
\cite<e.g.,>{WascherG96}.  In addition, rounding heuristics applied
to fractional LP-solutions often yield the optimal, integral
solution.  The combination of a very tight LP lower bound and good
upper bounding procedure results in very little search being performed
for almost all problem instances.  

This branch-and-price
LP-based approach does not seem to generalize straightforwardly to the MKP and
MCCP, in part due to the differences in the granularity of the
objective function. 
The objective function for bin packing counts the number of bins used.
On the other hand, the objective functions for the  MKP and MCCP sums the profits of
the items assigned to the containers. Thus, the number
of possible, distinct objective function values for the MKP and MCCP is much
larger than the number of distinct objective function values for
a bin packing problem of comparable size.
Therefore, even if we assume the existence of some formulation analogous
to the cutting-stock problem, 
it is not likely that rounding up the LP 
solution for the MKP, MCCP, or other problems with objective functions that are fine-grained compared to bin packing will result in an optimistic 
bound that is as accurate as for bin packing. This suggests that it may be difficult to develop a branch-and-price solver that is competitive for these problems.
On the other hand, since the granularity of the objective function for bin covering is the
same as that for bin packing, it is possible that a branch-and-price
approach could be applied to bin covering. However, we are unaware of
any such approach in the literature.

\section{Conclusions}
\label{sec:conclusions}

We studied bin completion, a branch-and-bound approach
for multi-container packing, knapsack, and covering problems.
While previous work focused on item-oriented, branch-and-bound strategies that assign one item at a time to containers, bin completion is a bin-oriented branch-and-bound algorithm that uses a dominance relationship between bin assignments to prune the search. We presented a general
framework for this approach, and showed its general utility and
applicability to multicontainer problems.
We proposed several extensions to bin completion that improve its
efficiency, including nogood pruning, nogood dominance pruning, variable and value ordering heuristics, and hybrid
incremental undominated completion generation.
We demonstrated the power of the bin completion approach by developing new, state-of-the-art algorithms for three fundamental multicontainer problems.
We showed that bin completion algorithms significantly outperform the
Mulknap \cite{Pisinger99} 
and MTM \cite{MartelloT81} and 
algorithms on
hard MKP instances.
We developed a new, state-of-the-art algorithm for the MCCP based on bin completion. We showed that by exploiting a more powerful dominance criterion, our new bin completion algorithms significantly outperform an early bin completion algorithm by Christofides, Mingozzi, and Toth \citeyear{ChristofidesMT79}.
We developed a new, state-of-the-art algorithm for bin covering based on bin completion, and showed that our bin completion algorithm significantly outperforms the item-oriented branch and bound algorithm by Labb\'{e}, Laporte, and Martello \citeyear{LabbeLM95}.
However, our results for bin packing were not competitive with the
state-of-the-art solver based on the cutting-stock approach.  We
showed that for all four of the problems studied, nogood dominance
pruning consistently improves upon the performance of bin completion
with nogood pruning.\footnote{We previously showed that nogood pruning
significantly improves performance over bin-completion without pruning in
\cite{Korf03}. This was also confirmed for the MCCP 
(Table \ref{tab:mccp-results1}), 
and for the MKP and bin covering in
preliminary experiments.}


While we have focused on four particular multicontainer problems in
this paper, there are many similar problems involving the assignment
of objects to multiple containers where similar dominance relationships between candidate bin assignments
can be exploited. Examples include the generalized assignment problem, 
a widely studied generalization of the MKP with many applications where the weight and profit of an item is a function of the container to which it is assigned
 \cite<e.g.,>{CattrysseV92,MartelloT90},
multiprocessor scheduling, which is equivalent to $k$-way number partitioning, \cite{DellAmicoM95}, and the segregated storage problem \cite{Neebe87,EvansT93}. In addition, there are variants of the problems we studied with additional constraints, such as the class-constrained multiple knapsack problem \cite{ShachnaiT01,ShachnaiT01a,KellererPP04} which has applications in multimedia file storage. 
Exploiting powerful dominance
criteria in a bin completion framework appears to be a promising
future direction for such multicontainer problems.

One issue with bin completion
is that as the number of unique items
grows, the number of undominated bin assignments grows rapidly. 
While we showed that hybrid incremental branching can significantly alleviate this problem (Section \ref{sec:bin-covering-hybrid-incremental}), a drawback is that it limits the 
utility of value-ordering heuristics 
that are applied to sort
undominated bin assignments {\em after} they are generated. 
Thus, an algorithm that generates undominated assignments in an order that conforms to a desired heuristic value ordering, rather than relying on sorting the assignments after they are all generated, is an area for future work.



\section*{Acknowledgments}
Thanks to Gleb Belov for running his solver on some of our
bin packing test instances. The anonymous reviewers provided many helpful comments and suggestions that improved this paper.
This research was supported by NSF under grant No. EIA-0113313, and 
by the Jet Propulsion Laboratory, California Institute of Technology, under a contract with the National Aeronautics and Space Administration.


\bibliography{multicontainer-journal,alex}
\bibliographystyle{theapa}

\end{document}